\documentclass[12pt]{iopart}

\usepackage{iopams}
\usepackage{rotating}
\usepackage{soul}
\usepackage{iopams}

\expandafter\let\csname equation*\endcsname\relax

\expandafter\let\csname endequation*\endcsname\relax

\usepackage{amsmath}
\usepackage{amssymb}
\usepackage{graphics}
\usepackage{graphicx}
\usepackage{oldlfont}
\usepackage{newlfont}
\usepackage{amsfonts}
\usepackage{ulem}
\usepackage{multirow}
\usepackage{url}
\usepackage{comment}
 \usepackage{tikz}
\usetikzlibrary{fadings,shapes.arrows,shadows,decorations.pathreplacing,angles,quotes}
\usepackage{xparse}
\usepackage{lipsum}

 \newcommand{\R}{{\mathbb R}}
   \usepackage{amsthm}
 \theoremstyle{definition}

\newcommand{\N}{\mathcal{N}}

\newcommand{\kk}{\mathbf k}

\newcommand{\q}{\mathbf q}
\newcommand{\vu}{\mathbf u}
\newcommand{\y}{\mathbf y}
\newcommand{\x}{\mathbf x}

\newcommand{\Transp}{\mathsf{T}}

\newcommand{\Bs}[1]{\boldsymbol{#1}}

\def\GP{\mathcal{GP}}
\newcommand{\GPd}[2]{\GP \left(#1,\,\, #2 \right)}

\usepackage[english]{babel}
\usepackage[utf8x]{inputenc}
\usepackage[T1]{fontenc}
\usepackage{xcolor}

\begin{document}

\title{Probabilistic approach to limited-data computed tomography reconstruction}

\author{Zenith Purisha$^1$, Carl Jidling$^2$, Niklas Wahlstr{\"o}m$^2$, Thomas B. Sch{\"o}n$^2$, Simo S{\"a}rkk{\"a}$^1$}

\address{$^1$Department of Electrical Engineering and Automation, Aalto University, Finland\\%
    $^2$Department of Information Technology, Uppsala University, Sweden\\[2ex]}
\ead{zenith.purisha@aalto.fi}
\vspace{10pt}
\begin{indented}
\item[] 2019
\end{indented}

\begin{abstract}
In this work, we consider the inverse problem of reconstructing the internal structure of an object from limited x-ray projections.
We use a Gaussian process prior to model the target function and estimate its (hyper)parameters from measured data. In contrast to other established methods, this comes with the advantage of not requiring any manual parameter tuning, which usually arises in classical regularization strategies. Our method uses a basis function expansion technique for the Gaussian process which significantly reduces the computational complexity and avoids the need for numerical integration. The approach also allows for reformulation of come classical regularization methods as Laplacian and Tikhonov regularization as Gaussian process regression, and hence provides an efficient algorithm and principled means for their parameter tuning. Results from simulated and real data indicate that this approach is less sensitive to streak artifacts as compared to the commonly used method of filtered backprojection.
\end{abstract}

\section{Introduction}
X-ray computed tomography (CT) imaging is a non-invasive method to recover the internal structure of an object by collecting projection data from multiple angles. The projection data is recorded by a detector array and it represents the attenuation of the x-rays which are transmitted through the object. Since the 1960s, CT has been used to a deluge of applications in medicine \cite{cormack1963representation,cormack1964representation,herman1979image,kuchment2013radon,national1996mathematics,shepp1978computerized} and industry \cite{akin2003computed,cartz1995nondestructive,de2014industrial}.

Currently, the so-called filtered back projection (FBP) is the reconstruction algorithm of choice because it is very fast \cite{avinash1988principles,buzug2008computed}. This method requires dense sampling of the projection data to obtain a satisfying image reconstruction. 
However, for some decades, the limited-data x-ray tomography problem has been a major concern in, for instance, the medical imaging community. The limited data case---also referred to as {\it sparse projections}---calls for a good solution for several important reasons, including:
\begin{itemize}
\item the needs to examine a patient by using low radiation doses to reduce the risk of malignancy or to {\it in vivo} samples to avoid the modification of the properties of living tissues,
\item geometric restrictions in the measurement setting make it difficult to acquire the complete data \cite{riis2018limited}, such as in {\it mammography} \cite{niklason1997digital,rantala2006wavelet,wu2003tomographic,zhang2006comparative} and electron imaging \cite{fanelli2008electron},
\item the high demand to obtain the data using short acquisition times and to avoid massive memory storage, and
\item the needs to avoid---or at least minimize the impact of---the moving artifacts during the acquisition.
\end{itemize}

Classical algorithms---such as FBP---fail to generate good image reconstruction when dense sampling is not possible and we only have access to limited data. The under-sampling of the projection data makes the image reconstruction (in classical terms) an {\it ill-posed} problem \cite{natterer1986mathematics}. In other words, the inverse problem is sensitive to measurement noise and modeling errors. Hence, alternative and more powerful methods are required. Statistical estimation methods play an important role in handling the ill-posedness of the problem by restating the inverse problem as a {\it well-posed extension} in a larger space of probability distributions \cite{kaipio2006statistical}. Over the years there have been a lot of work on tomographic reconstruction from limited data using statistical methods (see, e.g.,  \cite{rantala2006wavelet,bouman1996unified,haario2017shape,kolehmainen2003statistical,siltanen2003statistical,sauer1994bayesian}). %

In the statistical approach, incorporation of {\it a priori} knowledge is a crucial part in improving the quality of the image reconstructed from limited projection data. That can be viewed as an equivalent of the regularization parameter in classical regularization methods. However, statistical methods, unlike classical regularization methods, also provide a principled means to estimate the parameters of the prior (i.e., the hyperparameters) which corresponds to automatic tuning of regularization parameters.

In our work we build the statistical model by using a Gaussian process model \cite{Rasmussen2006} with a hierarchical prior in which the (hyper)parameters in the prior become part of the inference problem. As this kind of hierarchical prior can be seen as an instance of a Gaussian process (GP) regression model, the computational methods developed for GP regression in machine learning context \cite{Rasmussen2006} become applicable. It is worth noting that some works on employing GP methods for tomographic problems have also appeared before. An iterative algorithm to compute a maximum likelihood point in which the prior information is represented by GP is introduced in \cite{tarantola2005inverse}. In \cite{hendriks2018implementation,jidling2018probabilistic}, tomographic reconstruction using GPs to model the strain field from neutron Bragg-edge measurements has been studied. Tomographic inversion using GP for plasma fusion and soft x-ray tomography have been done in \cite{li2013bayesian,svensson2011non}. Nevertheless, the proposed approach is different from the existing work. 

Our aim is to employ a hierarchical Gaussian process regression model to reconstruct the x-ray tomographic image from limited projection data. Due to the measurement model involving line integral computations, the direct GP approach does not allow for closed form expressions. The first contribution of this article is to overcome this issue by employing the basis function expansion method proposed in \cite{SolinSarkka2015}, which makes the line integral computations tractable as it detaches the integrals from the model parameters. This approach can be directly used for common GP regression covariance functions such as Mat\'ern or squared exponential. The second contribution of this article is to point out that the we can also reformulate classical regularization, in particular Laplacian and Tikhonov regularization, as Gaussian process regression where only the spectral density of the process (although not the covariance function itself) is well defined. As the basis function expansion only requires the availability of the spectral density, we can build a hierarchical model off a classical regularization model as well and have a principles means to tune the regularization parameters. Finally, the third contribution is to present methods for hyperparameter estimation that arise from the machine learning literature and apply the methodology to the tomographic reconstruction problem. In particular, 
the proposed methods are applied to simulated 2D chest phantom data available in \textsc{Matlab} and real carved cheese data measured with $\mu$CT system. The results show that the reconstruction images created using the proposed GP method outperforms the FBP reconstructions in terms of image quality measured as relative error and as peak signal to noise ratio.

\section{Constructing the model}
\subsection{The tomographic measurement data}

Consider a physical domain $\Omega \subset \R^2$ and an attenuation function $f:\Omega\rightarrow\R$. The x-rays travel through $\Omega$ along straight lines and we assume that the initial intensity (photons) of the x-ray is $I_0$ and the exiting x-ray intensity is $I_d$. If we denote a ray through the object as function $s \mapsto (x_1(s),x_2(s))$ Then the formula for the intensity loss of the x-ray within a small distance $ds$ is given as: 

\begin{equation}\label{calibration1}
\frac{dI(s)}{I(s)}= -f(x_1(s),x_2(s)) ds,
\end{equation}
and by integrating both sides of \eqref{calibration1}, the following relationship is obtained
\begin{equation}\label{calibration2}
\int_{-R}^{R} f(x_1(s),x_2(s)) ds = \log\frac{I_0}{I_d},
\end{equation}
where $R$ is the radius of the object or area being examined.

\begin{figure}
\centering
\setlength{\unitlength}{0.75mm}
\begin{picture}(100,80)
\put(0,-65){\includegraphics[width=10cm]{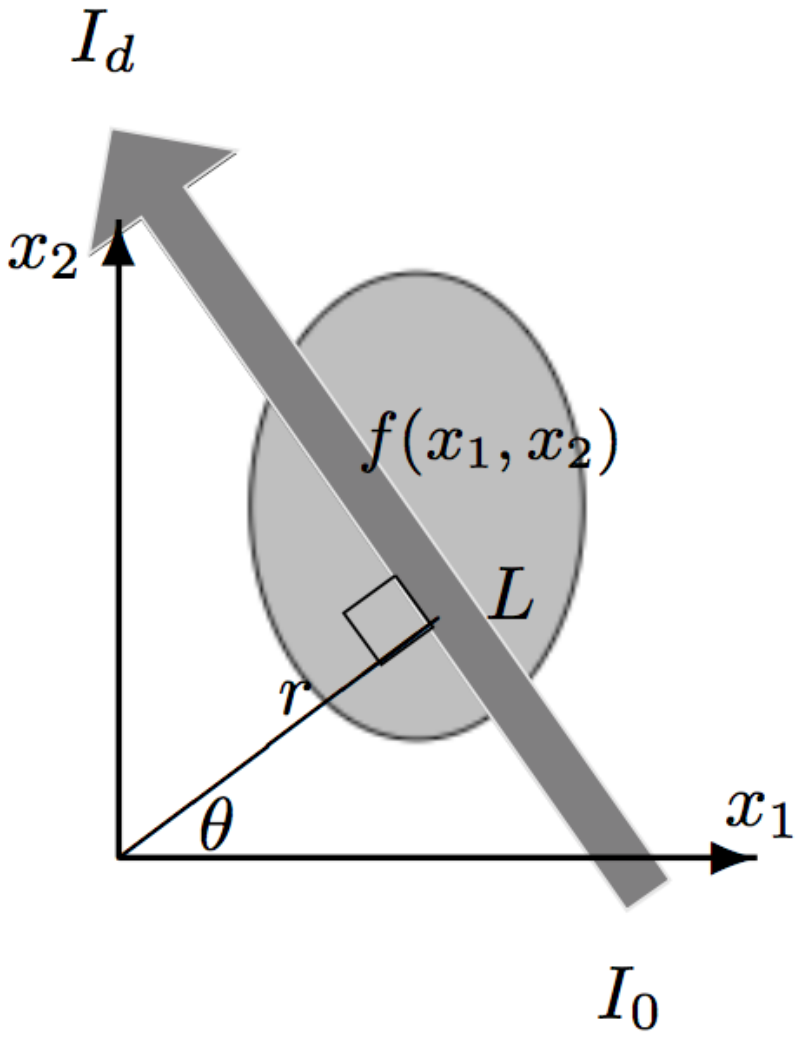}}
\end{picture}
\caption{An illustration of the Radon transform. It maps the object $f$ on the $(x_1,x_2)$-domain into $f$ on the $(r,\theta)$ domain. The measurement data is collected from the intensities $I_d$ of x-rays for all lines L through the object $f(x_1,x_2)$ and from different angles of view. }\label{Radon transform}
\end{figure}

In x-ray tomographic imaging, the aim is to reconstruct $f$ using measurement data collected from the intensities $I_d$ of x-rays for all lines through the object taken from different angles of view. 
The problem can be expressed using the Radon transform, which can be expressed as
\begin{equation}\label{Measurement Model}
\mathcal{R} f(r,\theta) = \int f(x_1,x_2) d\x_L,
\end{equation}
where $d\x_L$ denotes the $1$-dimensional Lebesgue measure along the line defined by $L=\{(x_1,x_2) \in \R^2 : x_1\cos \theta + x_2\sin\theta = r \}$, where $\theta\in[0,\pi)$ is the angle and $r\in\R$ is the distance of $L$ from the origin as shown in Figure~\ref{Radon transform}.

The parametrization of the straight line $L$ with respect to the arc length $s$ can be written as:
\begin{equation}
\begin{split}
  x_1(s,\theta,r) &= r \, \cos(\theta) - s \, \sin(\theta), \\
  x_2(s,\theta,r) &= r \, \sin(\theta) + s \, \cos(\theta). \\
\end{split}
\label{eq:xystr}
\end{equation}
In this work, the object is placed inside a circular disk with radius $R$. Then, as a function of $r$ and $\theta$ the line integral in (\ref{Measurement Model}) can be written as
\begin{equation}\label{Radon}
\begin{split}
     \mathcal{R} f(r,\theta) &= \int_{-R}^{R} 
    f(x_1(s,\theta,r),x_2(s,\theta,r)) \, ds \\
 &= \int_{-R}^R f(\x^0+s\hat{\vu}) ds,
\end{split}
\end{equation}
where 
\begin{align*}
	\x^0 = \begin{bmatrix} r\cos(\theta) & r\sin(\theta) \end{bmatrix}^\Transp, \qquad
	\hat{\vu} = \begin{bmatrix} -\sin(\theta) & \cos(\theta) \end{bmatrix}^\Transp.
\end{align*}

In a real x-ray tomography application,  the measurement is corrupted by at least two noise types: photons statistics and electronic noise. In x-ray imaging, a massive number of photons are usually recorded at each detector pixel. In such case, a Gaussian approximation for the attenuation data in $\eqref{calibration2}$ can be used \cite{bouman1992generalized,sachs19993d}. Recall that a logarithm of the intensity is involved in $\eqref{Radon}$, and so additive noise is a reasonable model for the electronic noise.

We collect a set of measurements as
\begin{equation}\label{noisy measurement}
    y_i =  \int_{-R}^R f(\x_i^0+s\hat{\vu}_i) ds + \varepsilon_i,
\end{equation}
where $i$ corresponds to the data point index. The corresponding inverse problem is given the noisy measurement data $\{y_i\}_{i=1}^n$ in \eqref{noisy measurement} to reconstruct the object $f$.

\subsection{Gaussian processes as functional priors}\label{functional priors}
A Gaussian process (GP) \cite{Rasmussen2006} can be viewed as a distribution over functions, where the function value in each point is treated as a Gaussian random variable.
To denote that the function $f$ is modeled as a GP, we formally write
\begin{align}\label{eq:fGP}
	f(\x) \sim \GPd{m(\x)}{k(\x,\x')}.
\end{align} 
The GP is uniquely specified by the \textit{mean function} $m(\x)=\mathbb{E}[f(\x)]$ and the \textit{covariance function} $k(\x,\x')=\mathbb{E}[(f(\x)-m(\x))(f(\x')-m(\x'))]$.
The mean function encodes our prior belief of the value of $f$ in any point. 
In lack of better knowledge it is common to pick $m(\x)=0$, a choice that we will stick to also in this paper.  
 
The covariance function on the other hand describes the covariance between two different function values $f(\x)$ and $f(\x')$.
The choice of covariance function is the most important part in the GP model, as it stipulates the properties assigned to $f$.
A few different options are discussed in Section \ref{sec:cov_func}.

\begin{figure}[h]
	\centering
    \includegraphics[width=\textwidth]{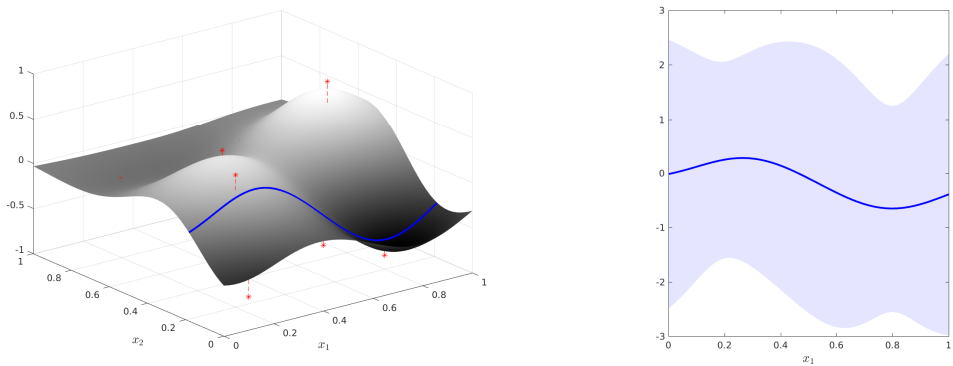}
	\caption{Left: GP prediction (shaded surface) obtained from the measurements (red stars, also indicated by their deviation from the prediction). Right: slice plot of the blue line in the left figure, including the $95\%$ credibility region.}
	\label{fig:se_ill_post}
\end{figure}

As data is collected our belief about $f$ is updated.
The aim of regression is to predict the function value $f(\x_*)$ at an unseen test point $\x_*$ by conditioning on the seen data.
Consider direct function measurements on the form 
\begin{equation}
	y_i=f(\x_i)+\varepsilon_i,
\end{equation}
where $\varepsilon_i$ is independent and identically distributed (iid) Gaussian noise with variance $\sigma^2$, that is, $\varepsilon_i\sim\N(0,\sigma^2)$. 
Let the measurements be stored in the vector $\y$.
Then the mean value and the variance of the predictive distribution $p(f(\x_*) \mid \y)$ are given by \cite{Rasmussen2006}
\begin{subequations}\label{eq:GPreg}
	\begin{align} 
	\mathbb{E}[f(\x_*) \mid \y] &=
	\kk_*^\Transp (K +\sigma^2I)^{-1}\y, \\
	\mathbb{V}[f(\x_*) \mid \y] &=
	k(\x_*,\x_*)-
	\kk_*^\Transp
	(K +\sigma^2I)^{-1}\kk_*.
	\end{align}
\end{subequations}
Here the vector $\kk_*$ contains the covariances between $f(\x_*)$ and each measurement while the matrix $K$ contains the covariance between all measurements, such that
\begin{subequations}
\begin{align}
	(\kk_*)_i &= k(\x_i,\x_*), \\
	K_{ij} &= k(\x_i,\x_j).
\end{align}
\end{subequations}
An example of GP regression for a two-dimensional input is given in Figure~\ref{fig:se_ill_post}. 
The red stars indicate the measurements, while the shaded surface is the GP prediction. 
The blue line highlights a slice of the plot that is shown explicitly to the right, including the $95\%$ credibility region.

\subsection{The Gaussian process for x-ray tomography}\label{GP Xray}

In this section, we show how to apply the functional priors presented in Section~\ref{functional priors} to x-ray tomography application. Since the x-ray measurements \eqref{Radon} are line integrals of the unknown function $f(\x)$, they are linear functionals of the Gaussian process. Hence, we can define a linear functional $\mathcal{H}_{\x,i}$ as follows:
\begin{align}
	\mathcal{H}_{\x,i} f(\x) = \int_{-R}^R f(\x^0_i+s\hat{\vu}_i) ds.
\end{align} 
and thus the GP regression problem becomes
\begin{subequations}
    \label{eq:invprob}
    \begin{align} 
	    f(\x) &\sim \GPd{m(\x)}{k(\x,\x')}, \\
	    y_i &= \mathcal{H}_{\x,i} f(\x)+\varepsilon_i.
    \end{align} 
\end{subequations}
As discussed, for example, in \cite{Sarkka:2011,SolinSarkka2015} the GP regression equations can be extended to this kind of models, which in this case leads to the following:
\begin{subequations}\label{eq:GPreg}
	\begin{align} 
	\mathbb{E}[f(\x_*)|\y] &=
	\q_*^\Transp (K +\sigma^2I)^{-1}\y, \\
	\mathbb{V}[f(\x_*)|\y] &=
	k(\x_*,\x_*)-
	\q_*^\Transp
	(Q +\sigma^2I)^{-1}\q_*,
	\end{align}
\end{subequations}
where $\mathbf{y} = \begin{bmatrix} y_1 & \cdots & y_n \end{bmatrix}^\Transp$ and
\begin{subequations}
\begin{align}
	\label{eq:crosscov}
	(\q_*)_i &= \int_{-R}^R k(\x_i^0+s\hat{\vu}_i,\x_*) ds, \\
    \label{eq:Gram}
    Q_{ij} &= \int_{-R}^R\int_{-R}^R k(\x_i^0+s\hat{\vu}_i,\x_j^0+s'\hat{\vu}_j) ds ds'.
\end{align}
\end{subequations}
In general we can not expect closed form solutions to \eqref{eq:crosscov}--\eqref{eq:Gram} and numerical computations are then required. However, even with efficient numerical methods, the process of selecting the hyperparameters is tedious since the hyperparameters are in general not decoupled from the integrand and the integrals need to be computed repeatedly in several iterations. In this paper, we avoid this by using the basis function expansion that will be described in Section~\ref{sec:approx}.

\subsection{Squared exponential and Mat\'ern covariance functions}\label{sec:cov_func} 
An important modeling parameter in Gaussian process regression is the covariance function $k(\x,\x')$ which can be selected in various ways. Because the basis function expansion described in Section \ref{sec:approx} requires the covariance function to be  \textit{stationary}, we here limit our discussion to covariance functions of this form. \textit{Stationarity} means that $k(\x,\x')=k(\Bs{\mathrm{r}})$ where $\Bs{\mathrm{r}}=\x-\x'$, so the covariance only depends on the distance between the input points. In that case we can also work with the spectral density, which is the Fourier transform of the stationary covariance function
\begin{equation} \label{eq:fourier}
S(\Bs{\omega}) =
\mathcal{F}[k]
= \int k(\Bs{\mathrm{r}})e^{-\text{i}\Bs{\omega}^\Transp\Bs{\mathrm{r}}}d\Bs{\mathrm{r}},
\end{equation}
where again $\Bs{\mathrm{r}}=\x-\x'$.

The perhaps most commonly used covariance function within the machine learning context \cite{Rasmussen2006} is the \textit{squared exponential} (SE) covariance function
\begin{align}\label{eq:SE}
	k_{\textrm{SE}}(\Bs{\mathrm{r}}) &= \sigma_f^2\exp\left[ -\frac{1}{2l^2}\|\Bs{\mathrm{r}}\|_2^2 \right],
\end{align} 
which has the following spectral density
\begin{align}\label{eq:SES}
    S_\textrm{SE}(\Bs{\omega})&=\sigma_f^2 (2\pi)^{d/2} l^d
    \exp\left[
    -\frac{l^2 \| \Bs{\omega} \|_2^2}{2}
    \right],
\end{align} 
where $d$ is the dimensionality of $\x$ (in our case $d=2$). The SE covariance function is characterized by the magnitude parameter $\sigma_f$ and the \textit{length scale} $l$. The squared exponential covariance function is popular due to its simplicity and ease of implementation. It corresponds to a process whose sample paths are infinitely many times differentiable and thus the functions modeled by it are very smooth.

Another common family of covariance functions is given by the Mat\'ern class
\begin{subequations}
\begin{align}
k_{\textrm{Matern}}(\Bs{\mathrm{r}})&=
\sigma_f^2
\frac{2^{1-\nu}}{\Gamma(\nu)}
\left(
\frac{\sqrt{2\nu}\|\Bs{\mathrm{r}}\|_2}{l}
\right)^\nu
K_\nu\left(
\frac{\sqrt{2\nu}\|\Bs{\mathrm{r}}\|_2}{l}
\right),
\\
S_{\textrm{Matern}}(\Bs{\omega})&=
\sigma_f^2\frac{2^d \pi^{d/2}\Gamma(\nu+d/2)(2\nu)^\nu}{\Gamma(\nu)l^{2\nu}}
\left(
\frac{2\nu}{l^2} + \| \Bs{\omega} \|_2^2
\right)^{-(\nu+d/2)},
\end{align}
\end{subequations}
where $K_\nu$ is a modified Bessel function \cite{Rasmussen2006}. 
The smoothness of the process is increased with the parameter $\nu$: in the limit $\nu\rightarrow\infty$ we recover the squared exponential covariance function.

Gaussian processes are also closely connected to classical spline smoothing \cite{kimeldorf1970} as well as other classical regularization methods \cite{kaipio2006statistical,mueller2012linear} for inverse problems. Although the construction of the corresponding covariance function is hard (or impossible), it is still possible to construct the corresponding spectral density in many cases. With these spectral densities and the basis function method of Section~\ref{sec:approx}, we can construct probabilistic versions of the classical regularization methods as discussed in the next section.

\subsection{Covariance functions arising from classical regularization}
Let us recall that a classical way to seek for solutions to inverse problems is via optimization of a functional of the form
\begin{equation}
  \mathcal{J}[f] = \frac{1}{2 \sigma^2} \sum_i (y_i - \mathcal{H}_{\x,i} f(\x))^2
  + \frac{1}{2 \sigma_f^2} \int | \mathcal{L} f(\x) |^2 \, d\x,
\end{equation}
where $\mathcal{L}$ is a linear operator. This is equivalent to a Gaussian process regression problem, where the covariance operator is formally chosen to be $\mathcal{K} = [\mathcal{L}^* \mathcal{L}]^{-1}$. In (classical) Tikhonov regularization we have $\mathcal{L} = \mathcal{I}$ (identity operator) which corresponds to penalizing the norm of the solution. Another option is to penalize the Laplacian which gives $\mathcal{L} = \nabla^2$.

Although the kernel of this covariance operator is ill-defined with the classical choices of $\mathcal{L}$ and thus it is not possible to form the corresponding covariance function, we can still compute the corresponding spectral density function by computing the Fourier transform \eqref{eq:fourier} of $\mathcal{L}^* \mathcal{L}$ and then inverting it to form the spectral density:
\begin{equation}
  S(\Bs{\omega}) = \frac{\sigma_f^2}{\mathcal{F}[\mathcal{L}^* \mathcal{L}]}.
\end{equation}
In particular, the minimum norm or (classical) Tikhonov regularization can be recovered by using a white noise prior which is given by the constant spectral density
\begin{equation} \label{eq:STikhonov}
  S_{\textrm{Tikhonov}}(\Bs{\omega}) = \sigma_f^2,
\end{equation}
where $\sigma_f$ is a scaling parameter. Another interesting case is the Laplacian operator based regularization which corresponds to
\begin{equation} \label{eq:SLaplacian}
  S_{\textrm{Laplacian}}(\Bs{\omega}) = \frac{\sigma_f^2}{\| \Bs{\omega} \|^4_2}.
\end{equation}
It is useful to note that the latter spectral density corresponds to a $l \to \infty$ limit of the Mat\'ern covariance function with $\nu + d/2 = 2$ and the white noise to $l \to 0$ in either the SE or the Mat\'ern covariance functions. The covariance functions corresponding to the above spectral densities would be degenerate, but this does not prevent us from using the spectral densities in the basis function expansion method described in Section~\ref{sec:approx} as the method only requires the availability of the spectral density.

\subsection{Basis function expansion}\label{sec:approx}
To overcome the computational hazard described in Section~\ref{GP Xray}, we consider the approximation method proposed in \cite{SolinSarkka2015}, which relies on the following truncated basis function expansion
\begin{align}
	\label{eq:BFE}
	k(\x,\x')\approx \sum_{i=1}^{m}S({\sqrt{\lambda}_i})\phi_i(\x)\phi_i(\x'),	
\end{align}
where $S$ denotes the spectral density of the covariance function, and $m$ is the truncation number.
The basis functions $\phi_i(\x)$ and eigenvalues ${\lambda}_i$ are obtained from the solution to the Laplace eigenvalue problem on the domain $\Omega$
\begin{align}\label{eq:lapl_eig}
\begin{cases}
\hspace{-4mm}
\begin{split}
-\Delta\phi_i(\x)&  ={\lambda}_i\phi_i(\x),  \\
\phi_i(\x)&         =0,         
\end{split}
\end{cases}
\quad
\begin{split}
\x&\in\Omega, \\
\x&\in\partial\Omega. 
\end{split}
\end{align}
In two dimensions with $\Omega=[-L_1,L_1]\times[-L_2,L_2]$ we introduce the positive integers $i_1\le m_1$ and $i_2\le m_2$.
The number of basis functions is then $m=m_1m_2$ and the solution to \eqref{eq:lapl_eig} is given by
\begin{subequations}
\begin{align}
\phi_i(\x)  &= \frac{1}{\sqrt{L_1L_2}}\sin\big(\varphi_{i_1}(x_1+L_1)\big)\sin\big(\varphi_{i_2}(x_2+L_2)\big) ,
\\
\lambda_i &= \varphi_{i_1}^2+\varphi_{i_2}^2, \quad \varphi_{i_1}=\frac{\pi i_1}{2L_1}, \quad \varphi_{i_2}=\frac{\pi i_2}{2L_2}, 
\end{align}
\end{subequations}
where $i=i_1+m_1(i_2-1)$.
Let us now build the vector ${\boldsymbol\phi}_*\in\R^{m\times1}$, the matrix $\Phi\in\R^{m\times M}$ and the diagonal matrix $\Lambda\in\R^{m\times m}$ as
\begin{subequations}
\begin{align}
	({\boldsymbol\phi}_*)_i&=\phi_i(\x_*), \\
    \label{eq:Phi_entry}
    \Phi_{ij} &= \int_{-R}^R \phi_i(\x^0_j+s\hat{\vu}_j)ds, \\
    \Lambda_{ii} &= S(\sqrt{\lambda_i}). 
\end{align}
\end{subequations}
The entries $\Phi_{ij}$ can be computed in closed form with details given in \ref{app:compdet}. 
Now we substitute $Q\approx\Phi^\Transp\Lambda\Phi$ and $\q_*\approx\Phi^\Transp\Lambda{\boldsymbol\phi}_*$ to obtain
\begin{subequations}\label{eq:pred_appr2}
\begin{align}
\mathbb{E}[f(\x_*) \mid \y]
&\approx{\boldsymbol\phi}_{*}^\Transp \Lambda \Phi (\Phi^\Transp \Lambda \Phi + \sigma^2 I)^{-1}\y,
\\
\mathbb{V}[f(\x_*) \mid \y]
&\approx {\boldsymbol\phi}_{*}^\Transp \Lambda {\boldsymbol\phi}_{*} -  {\boldsymbol\phi}_{*}^\Transp \Lambda \Phi (\Phi^\Transp \Lambda \Phi + \sigma^2I)^{-1} \Phi^\Transp \Lambda  {\boldsymbol\phi}_{*}.
\end{align}
\end{subequations}
When using the spectral densities corresponding to the classical regularization methods in \eqref{eq:STikhonov} and \eqref{eq:SLaplacian}, the mean equation reduces to the classical solution (on the given basis). However, also for the classical regularization methods we can compute the variance function which gives uncertainty estimate for the solution which in the classical formulation is not available. Furthermore, the hyperparameter estimation methods outlined in the next section provide principled means to estimate the parameters also in the classical regularization methods.

\section{Hyperparameter estimation}

In this section, we will consider some methods for estimating the \textit{hyperparameters}. The free parameters of the covariance function, for example, the parameters $\sigma_f$ and $l$ in the squared exponential covariance function, are together with the noise parameter $\sigma$ referred to as the hyperparameters of the model. In this work, we employ a Bayesian approach to estimate the hyperparameters, and comparisons with standard parameter estimation methods such as L-curve and cross-validation methods are given as well. 

\subsection{Posterior distribution of hyperparameters}

The marginal likelihood function corresponding to the model \eqref{eq:invprob} is given as
\begin{equation}\label{likelihood}
 p(\mathbf{y} \mid \sigma_f,l,\sigma)
 = \mathcal{N}(\mathbf{y} \mid \mathbf{0}, Q(\sigma_f,l) + \sigma^2 \, I),
\end{equation}
where $Q(\sigma_f,l)$ is defined by \eqref{eq:crosscov}. The posterior distribution of parameters can now be written as follows:

\begin{equation}\label{posterior}
    p(\sigma_f,l,\sigma \mid \mathbf{y}) \propto
    p(\mathbf{y} \mid \sigma_f,l,\sigma) 
    p(\sigma_f) p(l) p(\sigma),
\end{equation}
where non-informative priors are used: $p(\sigma_f)\propto \frac{1}{\sigma_f}$, $p(l)\propto \frac{1}{l}$ and $p(\sigma)\propto \frac{1}{\sigma}$. The logarithm of~\eqref{posterior} can be written as
\begin{equation}\label{logposterior}
    \log\,p(\sigma_f,l,\sigma \mid \mathbf{y}) =
    \text{const.} - \frac{1}{2} \, \log \det (Q + \sigma^2 I) 
    - \frac{1}{2} \mathbf{y}^\Transp (Q + \sigma^2 I)^{-1} \mathbf{y} 
    - \log \frac{1}{\sigma_f}
    - \log \frac{1}{l} - \log \frac{1}{\sigma}. 
\end{equation}
Given the posterior distribution we have a wide selection of methods from statistics to estimate the parameters. One approach is to compute the maximum a posteriori (MAP) estimate of the parameters by using, for example, gradient-based optimization methods \cite{Rasmussen2006}. However, using this kind of point estimate loses the uncertainty information of the hyperparameters and therefore in this article we use Markov chain Monte Carlo (MCMC) methods \cite{brooks2011handbook} which retain the information about the uncertainty in the final result. 

\subsection{Metropolis--Hastings sampling of hyperparameters}

As discussed in the previous section, the statistical formulation of the inverse problem gives a posterior distribution of the hyperparameters $\Bs{\varphi} = (\sigma_f,l,\sigma)$ as the solution rather than single estimates. The MCMC methods are capable of  generating samples from the distribution. The Monte Carlo samples can then be used for computing the mean, the variance, or some other statistics of the posterior distribution \cite{Gelman_et_al:2013}. In this work, we employ the Metropolis--Hastings algorithm to sample from the posterior distribution.

\subsection{The L-curve method}
One of the classical methods to obtain information about the optimum value for $\sigma$ is the L-curve method \cite{hansen1992analysis}, which operats by plotting the norm of the solution $\lVert f_{\sigma}(\x) \rVert _2$ versus the residual norm $\lVert \mathcal{H}_{\x,i} f_{\sigma}(\x) - y_i \rVert _2$. The associated L-curve is defined as the continous curve consisting of all the points $(\lVert \mathcal{H}_{\x,i} f_{\sigma}(\x) - y_i \rVert _2,\lVert f_{\sigma}(\x) \rVert _2)$ for $\sigma \in [0,\infty)$.

\subsection{Cross-validation}
As a comparison, we also consider to use methods of cross-validation (CV) for model selection. In $k$-fold CV, the data are partitioned into $k$ disjoint sets $\mathbf{y}_{j}$, and at each round $j$ of CV, the predictive likelihood of the set $\mathbf{y}_{j}$ is computed given the rest of the data $\mathbf{y}_{-j}$. These likelihoods are used to monitor the predictive performance of the model. This performance is used to estimate the generalization error, and it can be used to carry out model selection \cite{kohavi1995study,Rasmussen2006,vehtari2017practical}.

The Bayesian CV estimate of the predictive fit with given parameters $\Bs{\varphi}$ is
\begin{equation}
    \mbox{CV} = \sum_{j=1}^n
    \log p(\mathbf{y}_j \mid \mathbf{y}_{-j},\Bs{\varphi}),
\end{equation}
where $p(\mathbf{y}_j \mid \mathbf{y}_{-j},\Bs{\varphi})$ is the predictive likelihood of the data $\mathbf{y}_j$ given the rest of the data. The best parameter values with respect to CV can be computed by enumerating the possible parameter values and selecting the one which gives the best fit in terms of CV.

\section{Experimental results}\label{sec:expResults}
In this section, we present numerical results using the GP model for limited x-ray tomography problems. All the computations were implemented in \textsc{Matlab} 9.4 (R2018a) and performed on an Intel Core i5 at 2.3 GHz and CPU 8GB 2133MHz LPDDR3 memory.

For both simulated data (see Section~\ref{SimulatedData}) and real data (see Section~\ref{RealData}) we use $m=10^4$ basis functions in~\eqref{eq:BFE}. 
The measurements are obtained from the line integral of each x-ray over the attenuation coefficient of the measured objects. The measurements are taken for each direction (angle of view), and later they will be referred to as projections. The same number of rays in each direction is used.
The computation of the hyperparameters is carried out using the Metropolis--Hastings algorithms with $5\,000$ samples, and the first $1\,000$ samples are thrown away ({\it burn-in} period). 
The reconstruction is computed by taking the conditional mean of the object estimate. 

\subsection{Simulated data: 2D Chest phantom}\label{SimulatedData}
\begin{figure}
\begin{picture}(100,260)
\put(5,138){\includegraphics[width=6.3cm]{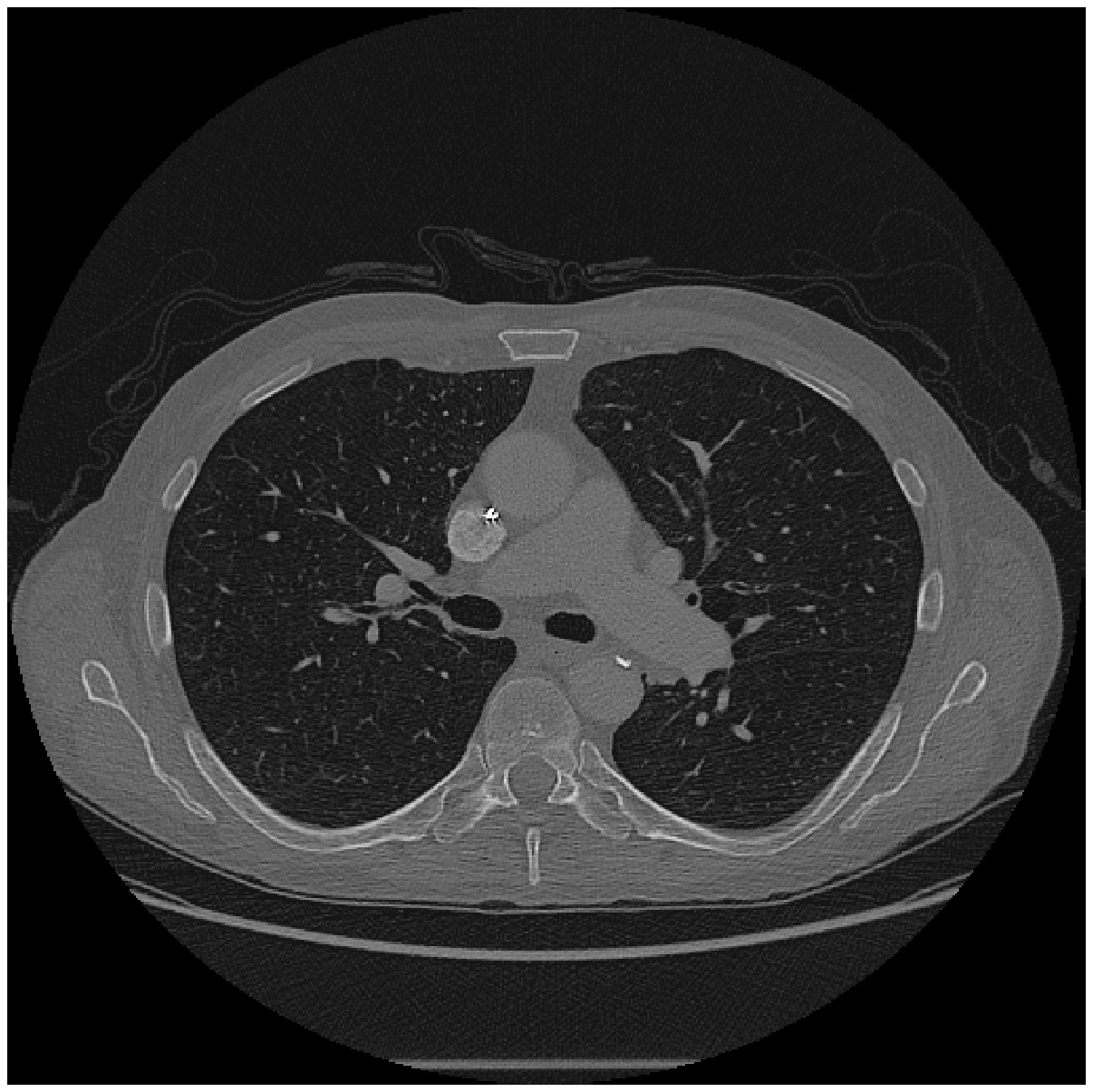}}
\put(158,138){\includegraphics[width=6.3cm]{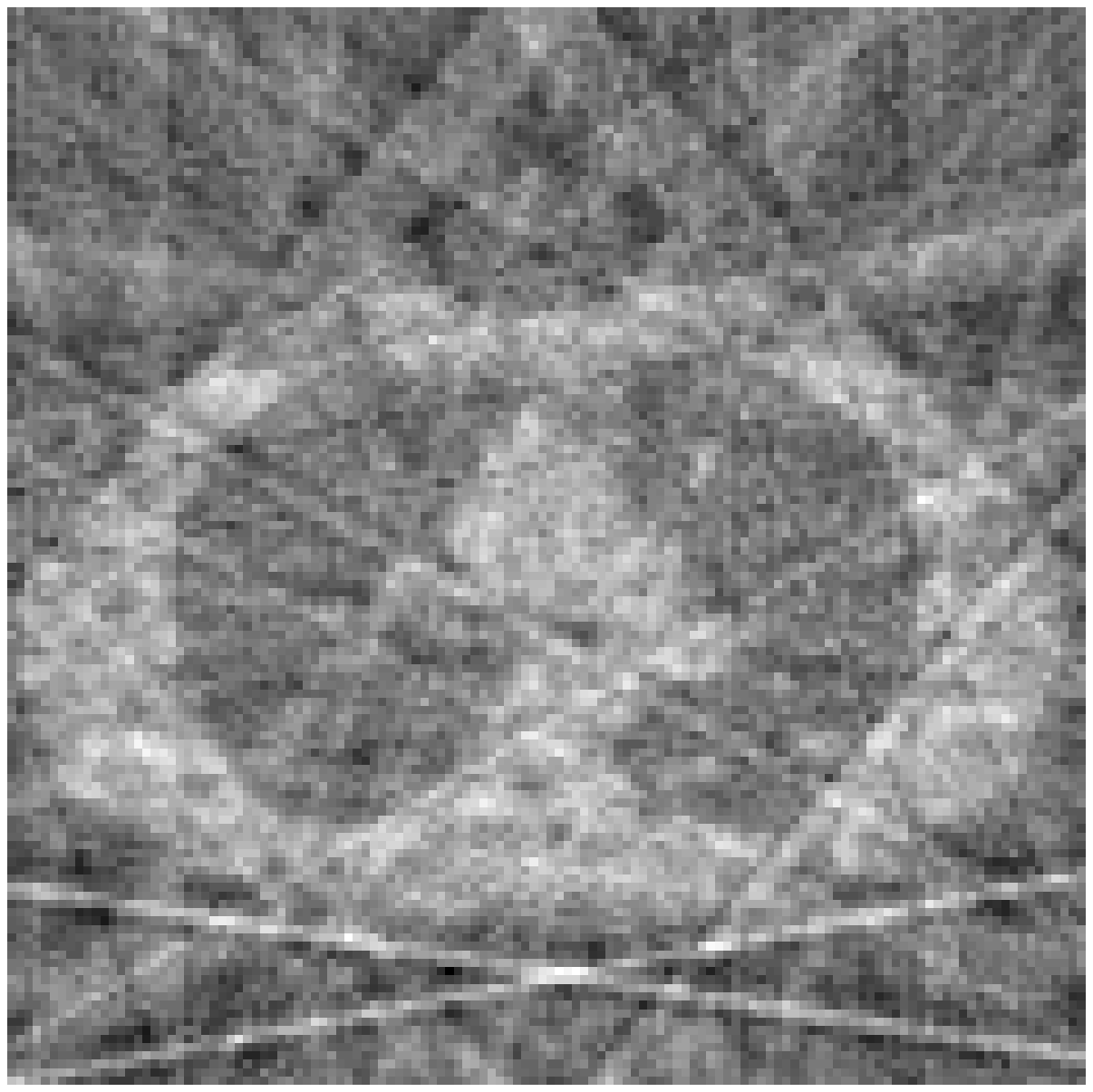}}
\put(306,138){\includegraphics[width=6.3cm]{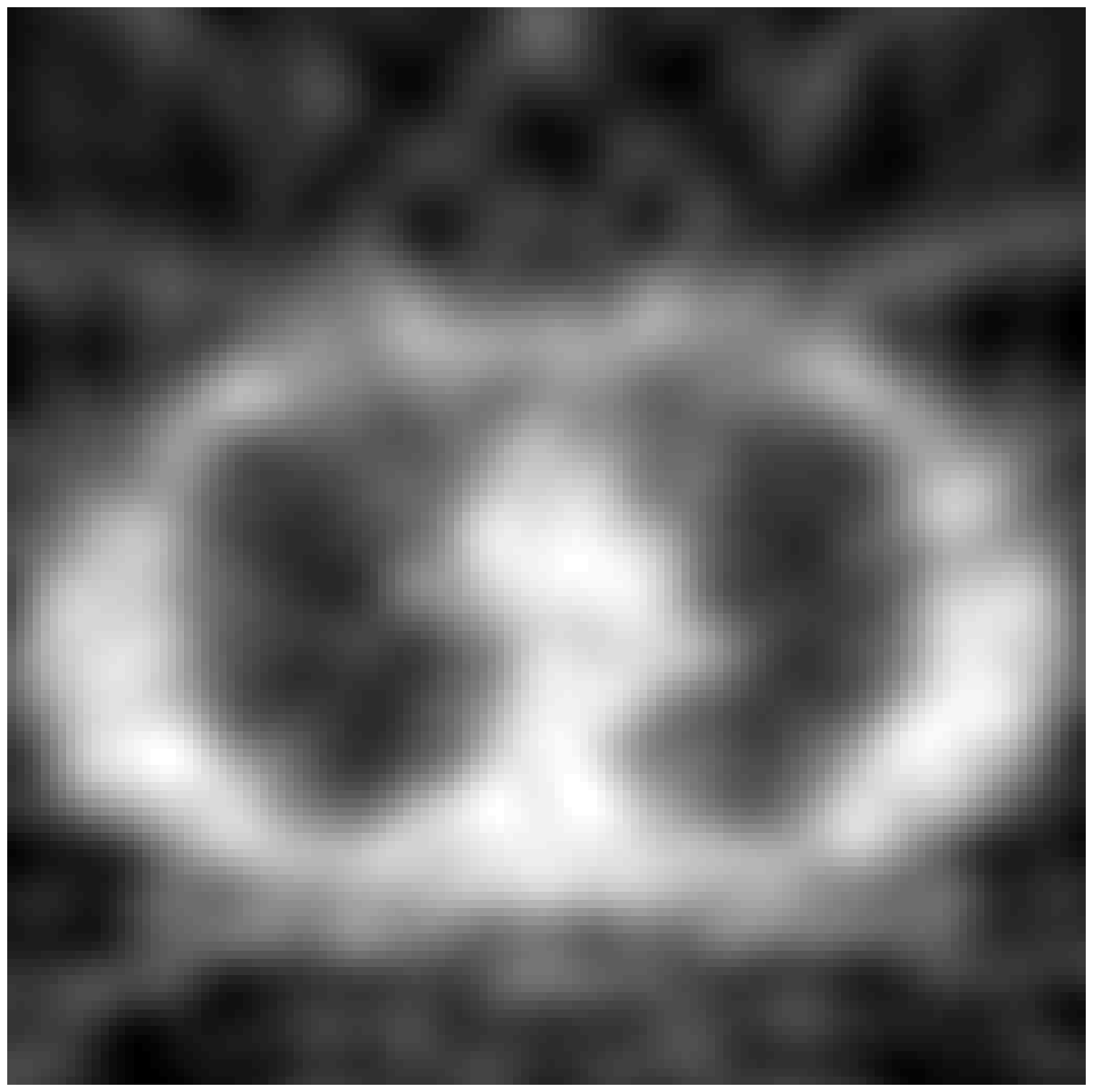}}
\put(5,3){\includegraphics[width=6.3cm]{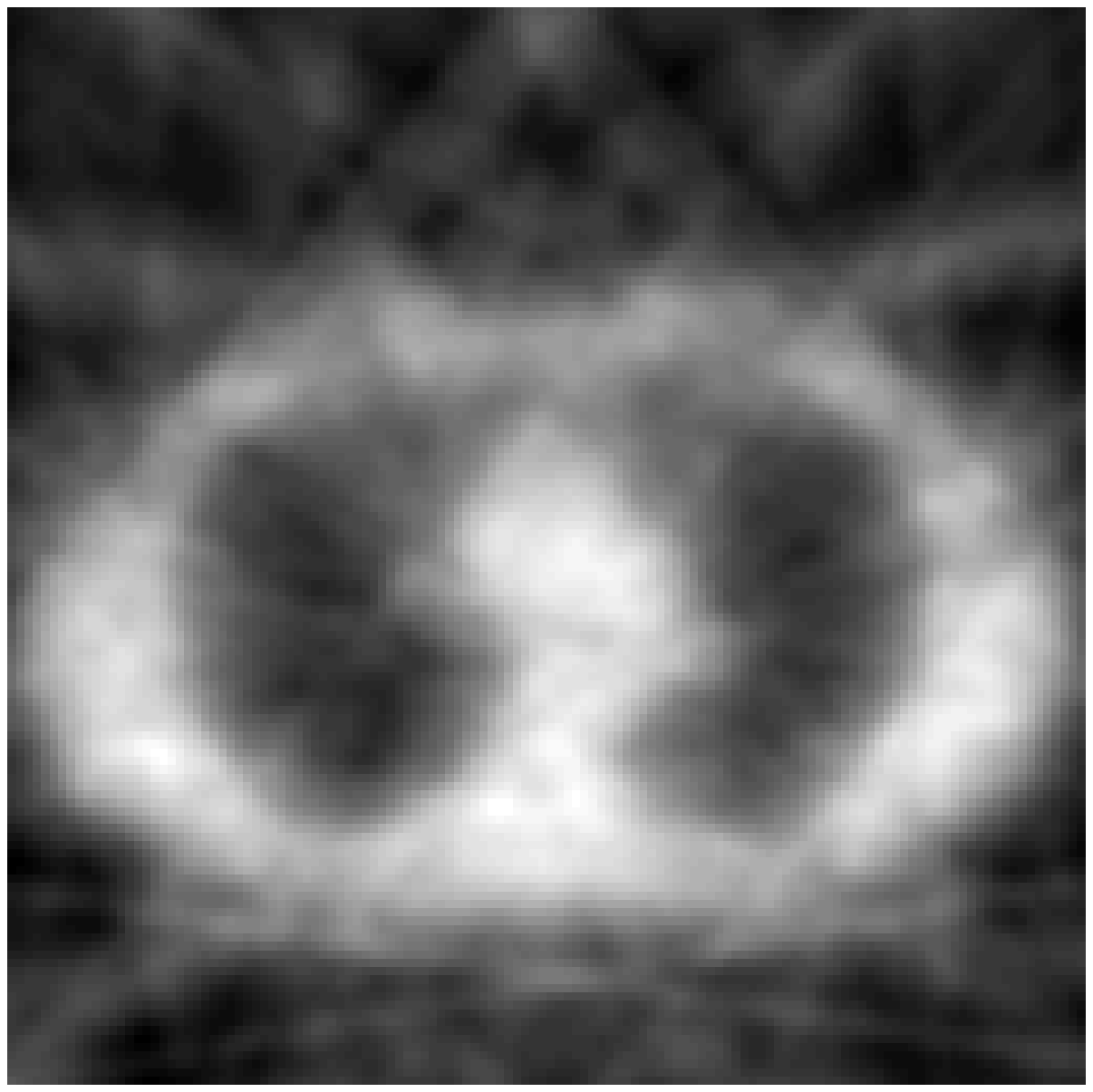}}
\put(158,3){\includegraphics[width=6.3cm]{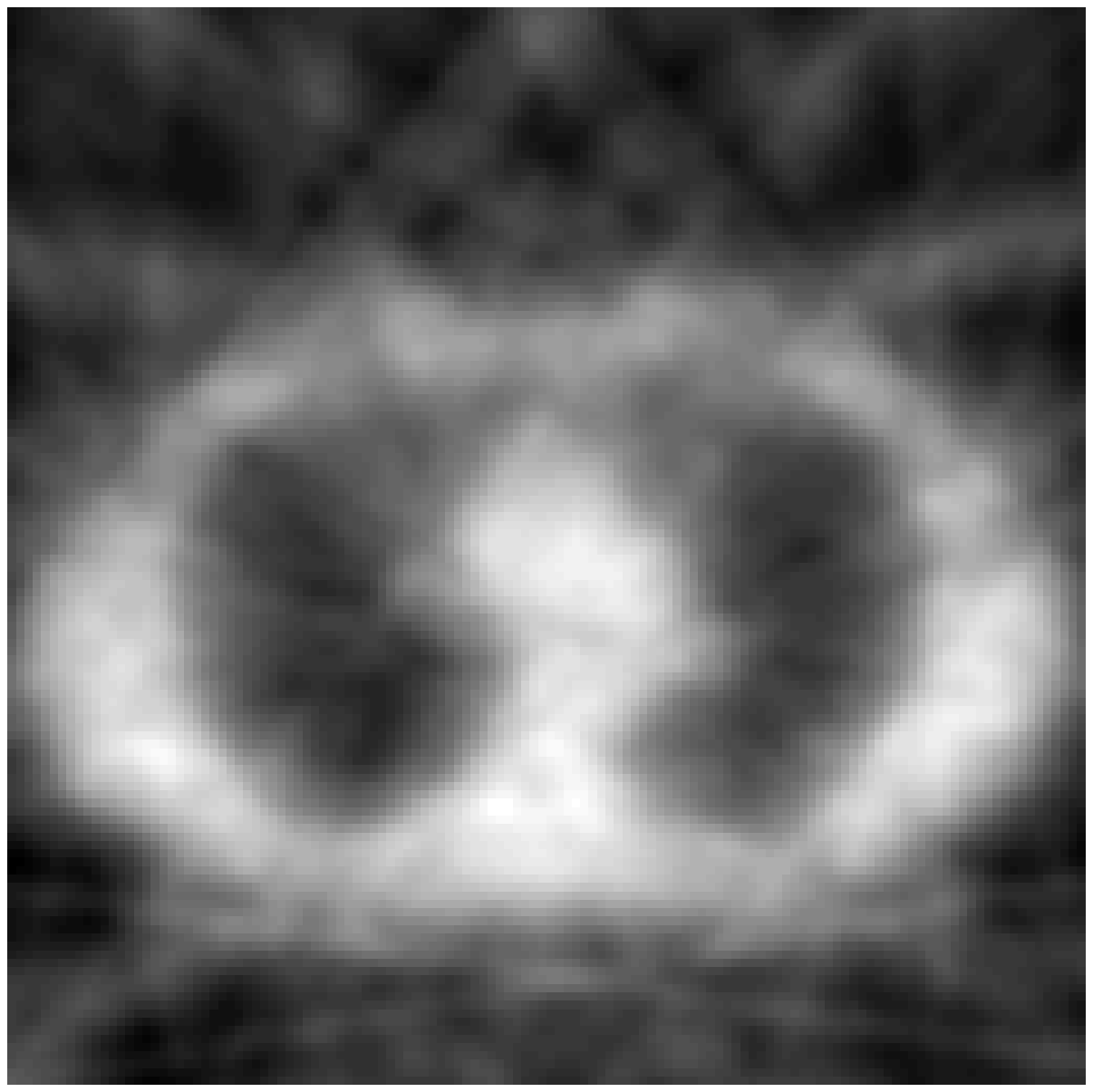}}
\put(306,3){\includegraphics[width=6.3cm]{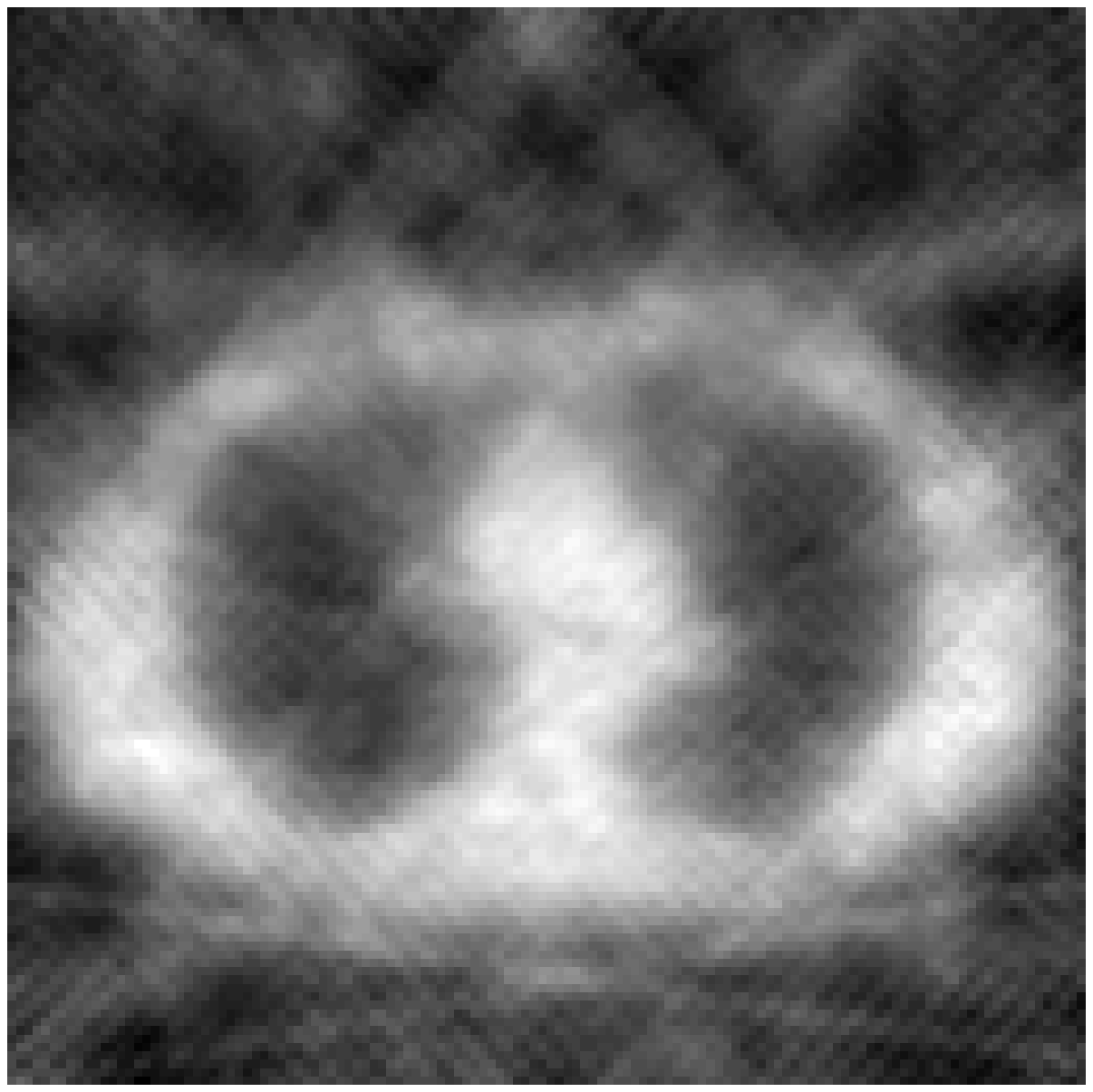}}
\put(87,137){(a)}
\put(240,137){(b)}
\put(393,137){(c)}
\put(87,0){(d)}
\put(240,0){(e)}
\put(393,0){(f)}
\end{picture}
\caption{(a) A ground truth of 2D chest phantom. (b) Filtered backprojection reconstruction (Ram--Lak filter) from 9 projections. (c) GP reconstruction using SE covariance, (d) GP reconstruction using Mat\'ern covariance, (e) GP reconstruction using Laplacian covariance, (f) GP reconstruction using Tikhonov covariance. The GP reconstructions are using 9 projections.}
\label{fig:ChestPhantomRec}
\end{figure}

As for the simulated data, we use one slice of \textsc{Matlab}'s 3D Chest dataset \cite{matlab_chest} as a ground truth, $f_{\text{true}}$, which is shown in Figure~\ref{fig:ChestPhantomRec}(a). The size of the phantom is $N \times N$, with $N = 128$. The black region indicates zero values and lighter regions indicate higher attenuation function values. 
The measurements (i.e. sinogram) of the chest phantom are computed using the {\tt radon} command in \textsc{Matlab} and corrupted by additive white Gaussian noise with zero mean and $0.1$ variance ($\sigma_{\text{true}} = 0.32$).

\begin{figure}
\begin{picture}(100,600)
\put(52,415){\includegraphics[width=4.25cm]{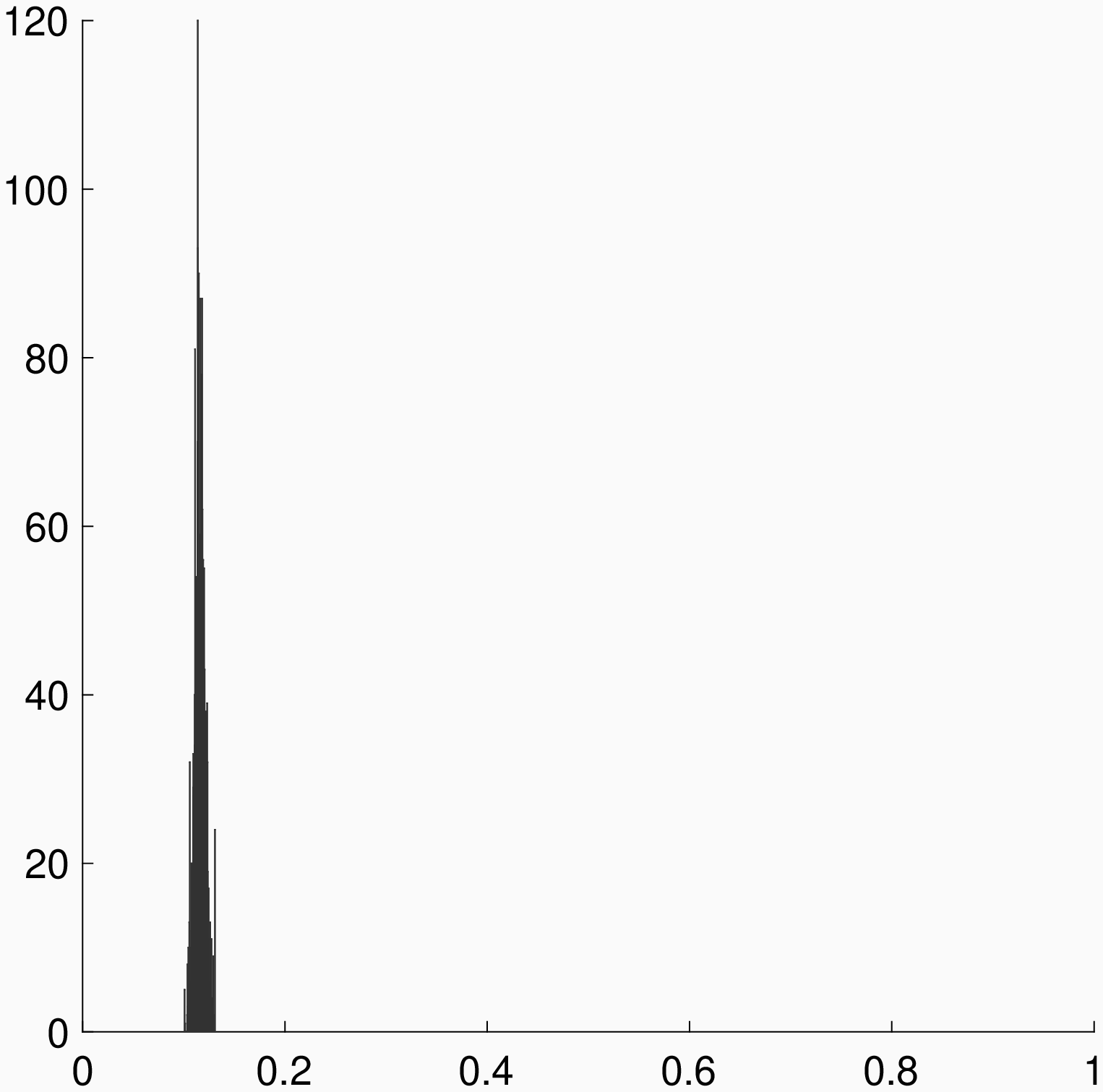}}
\put(190,415){\includegraphics[width=4.25cm]{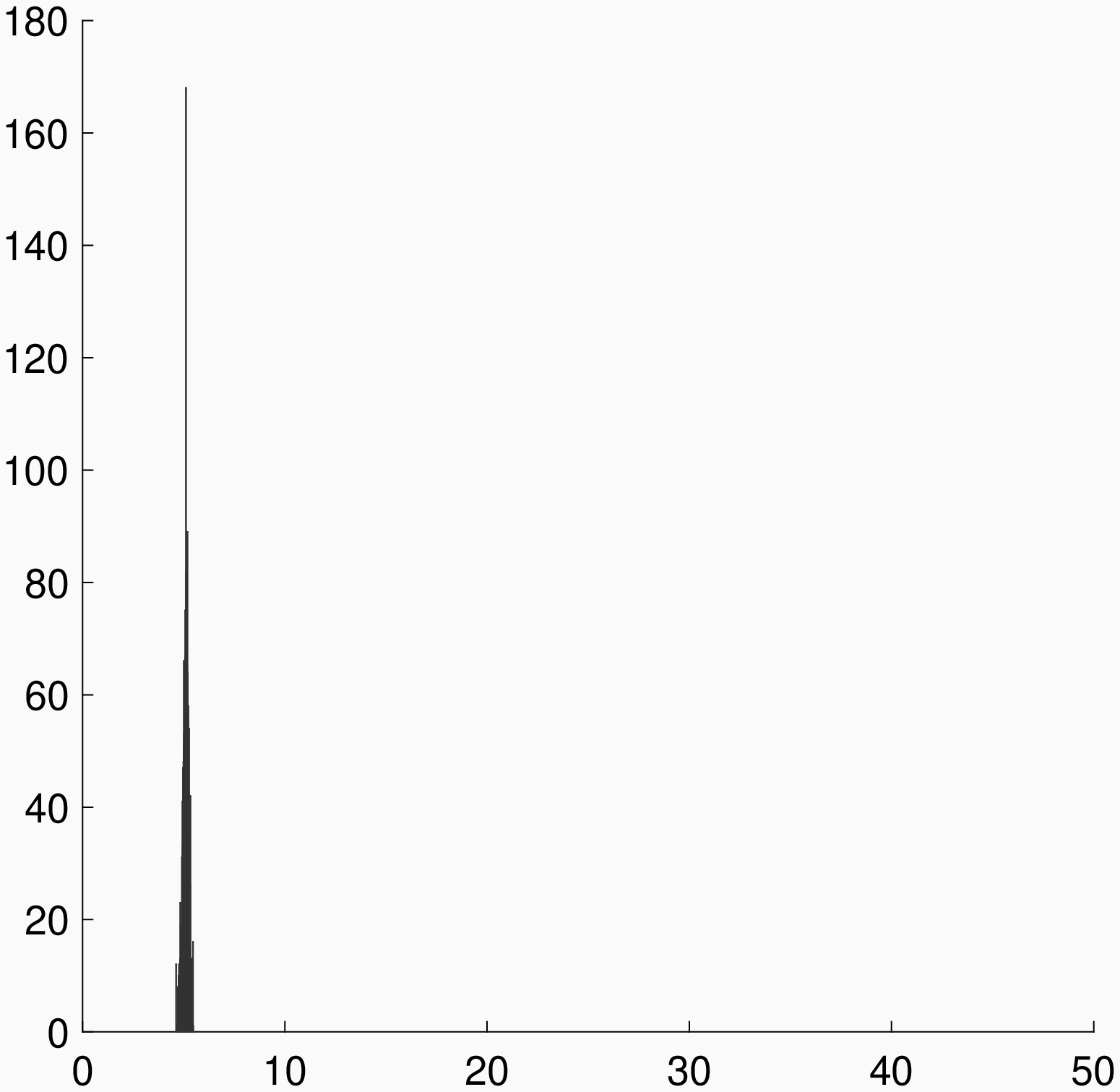}}
\put(326,415){\includegraphics[width=4.25cm]{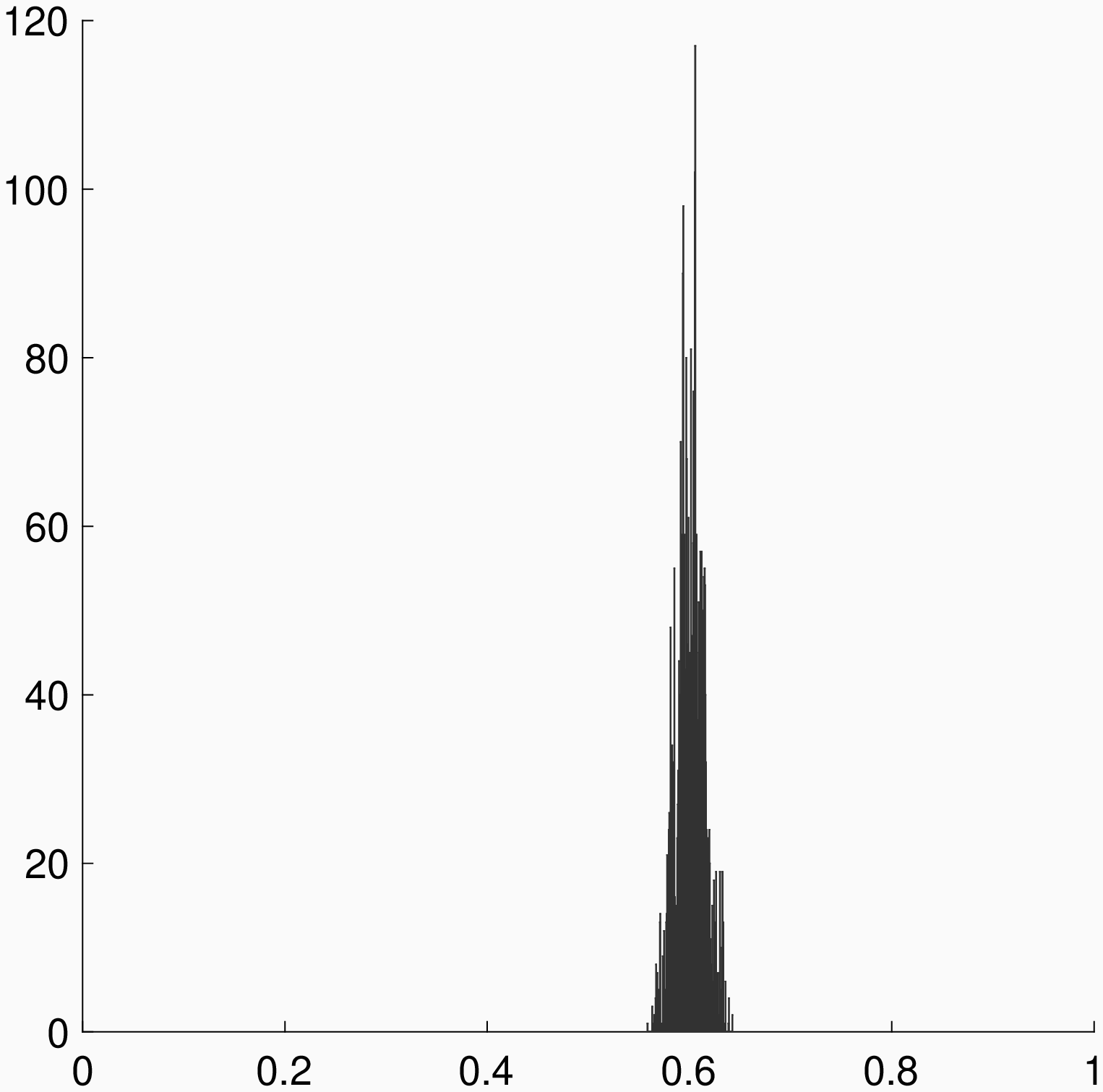}}
\put(52,265){\includegraphics[width=4.25cm]{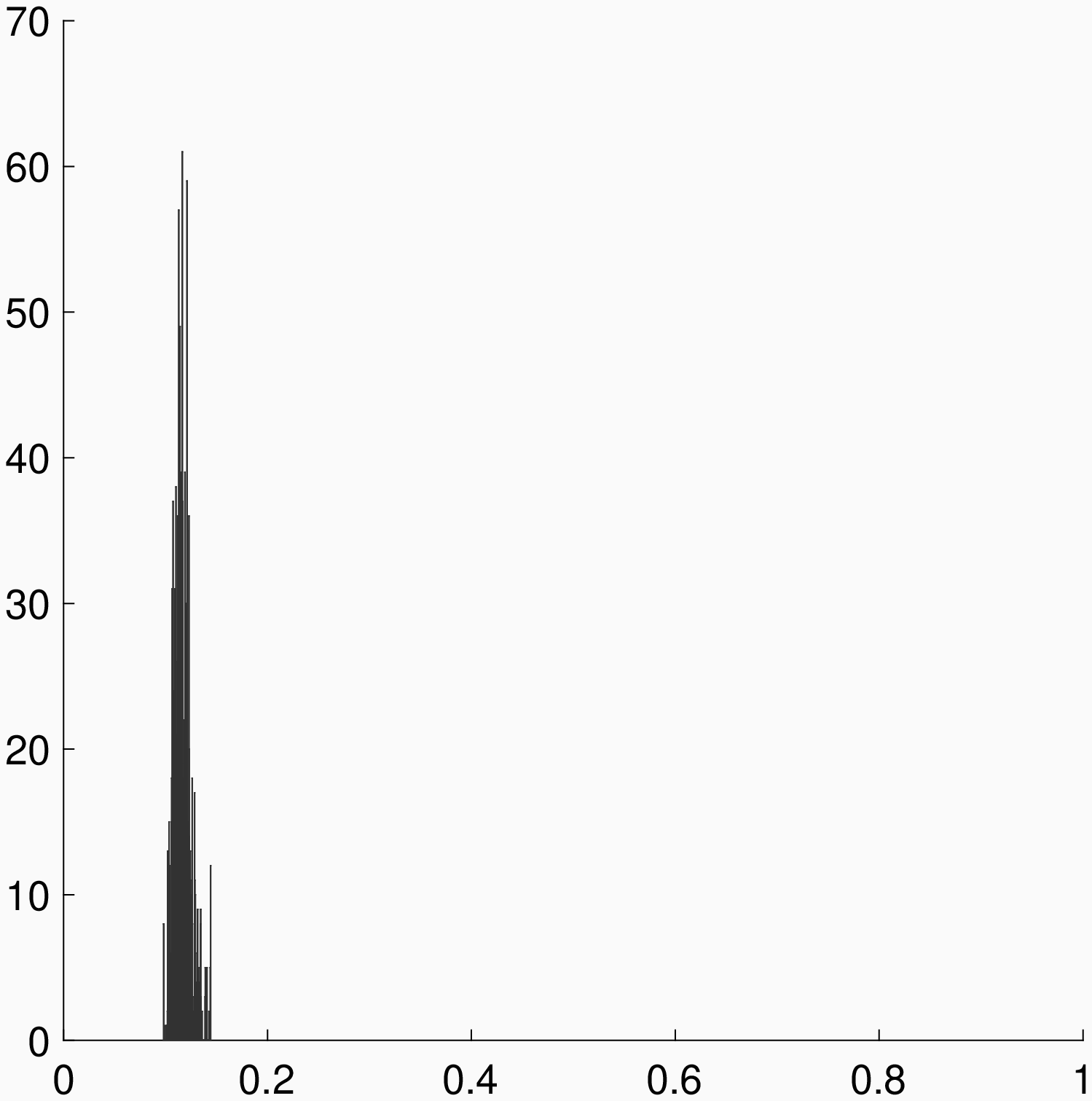}}
\put(190,265){\includegraphics[width=4.25cm]{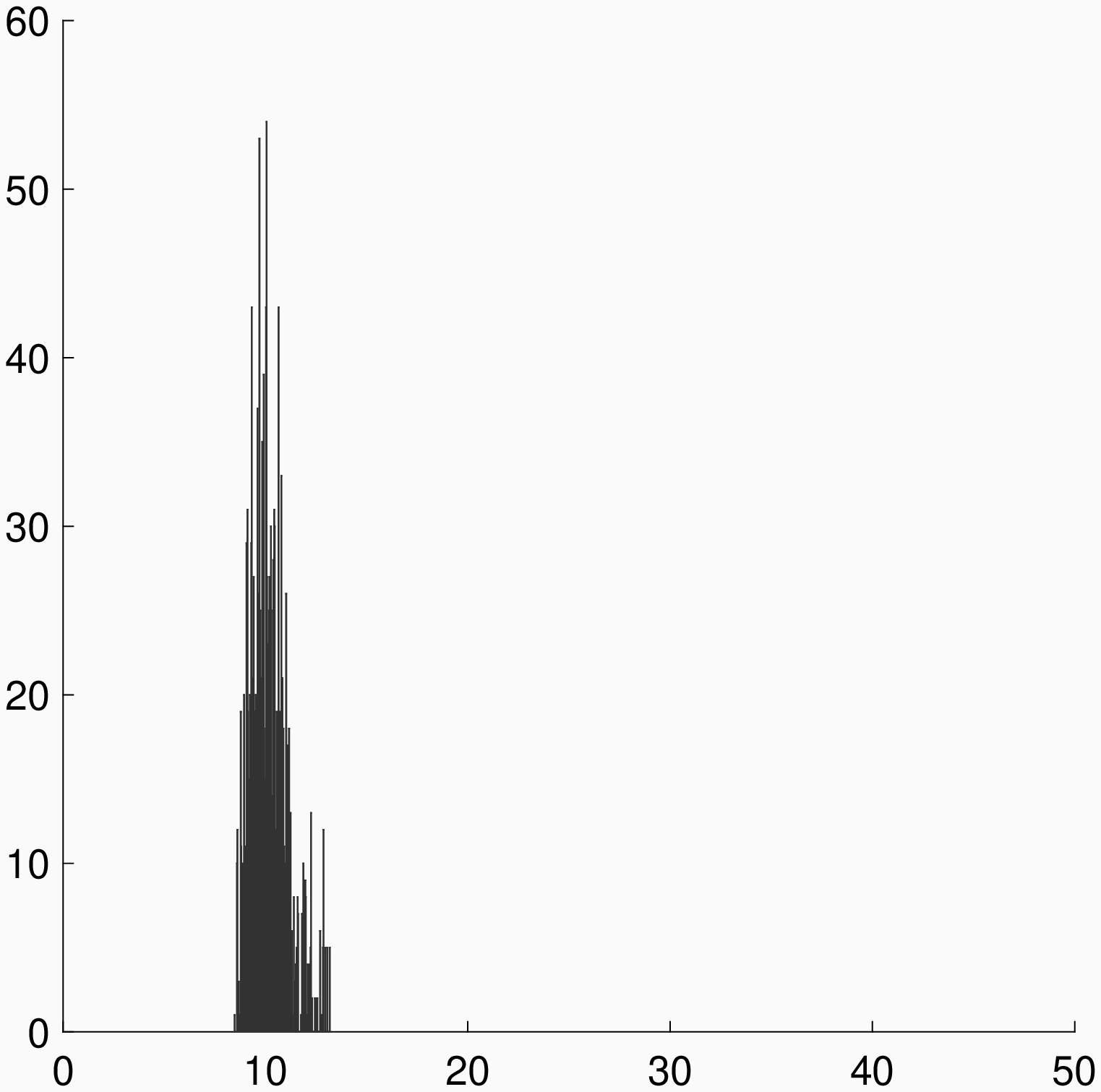}}
\put(326,265){\includegraphics[width=4.25cm]{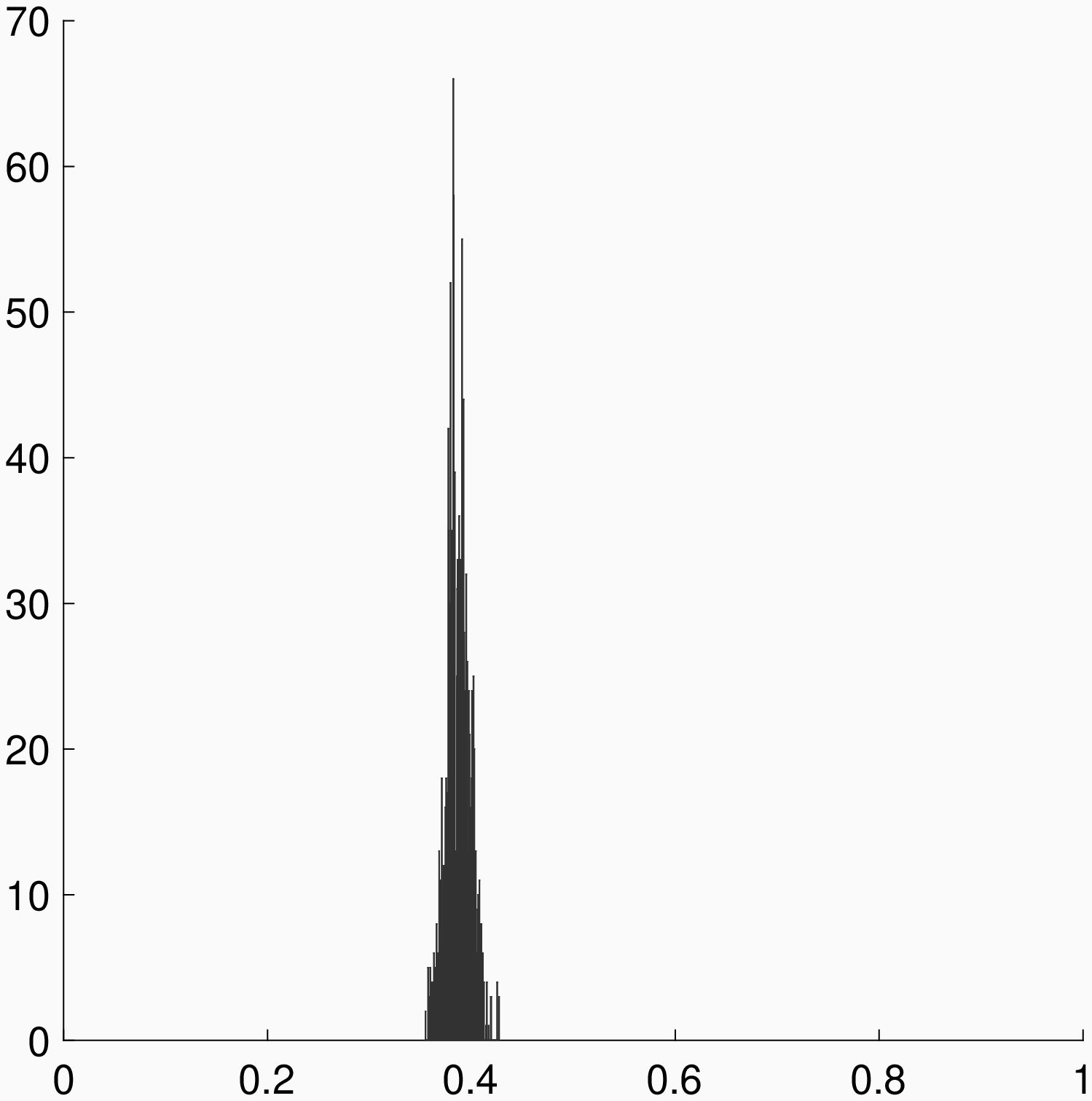}}
\put(52,125){\includegraphics[width=4.25cm]{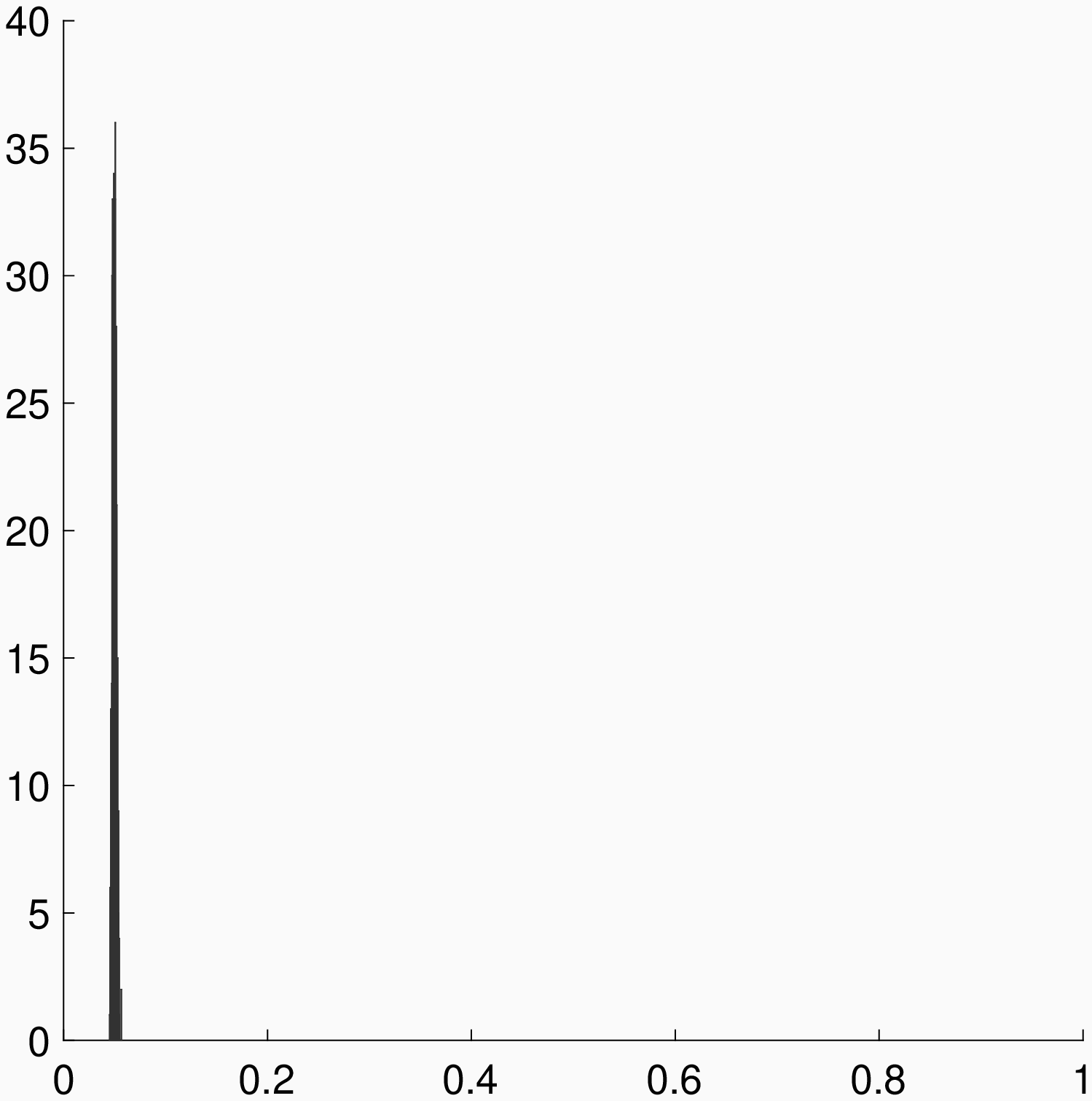}}
\put(185,125){\includegraphics[width=4.25cm]{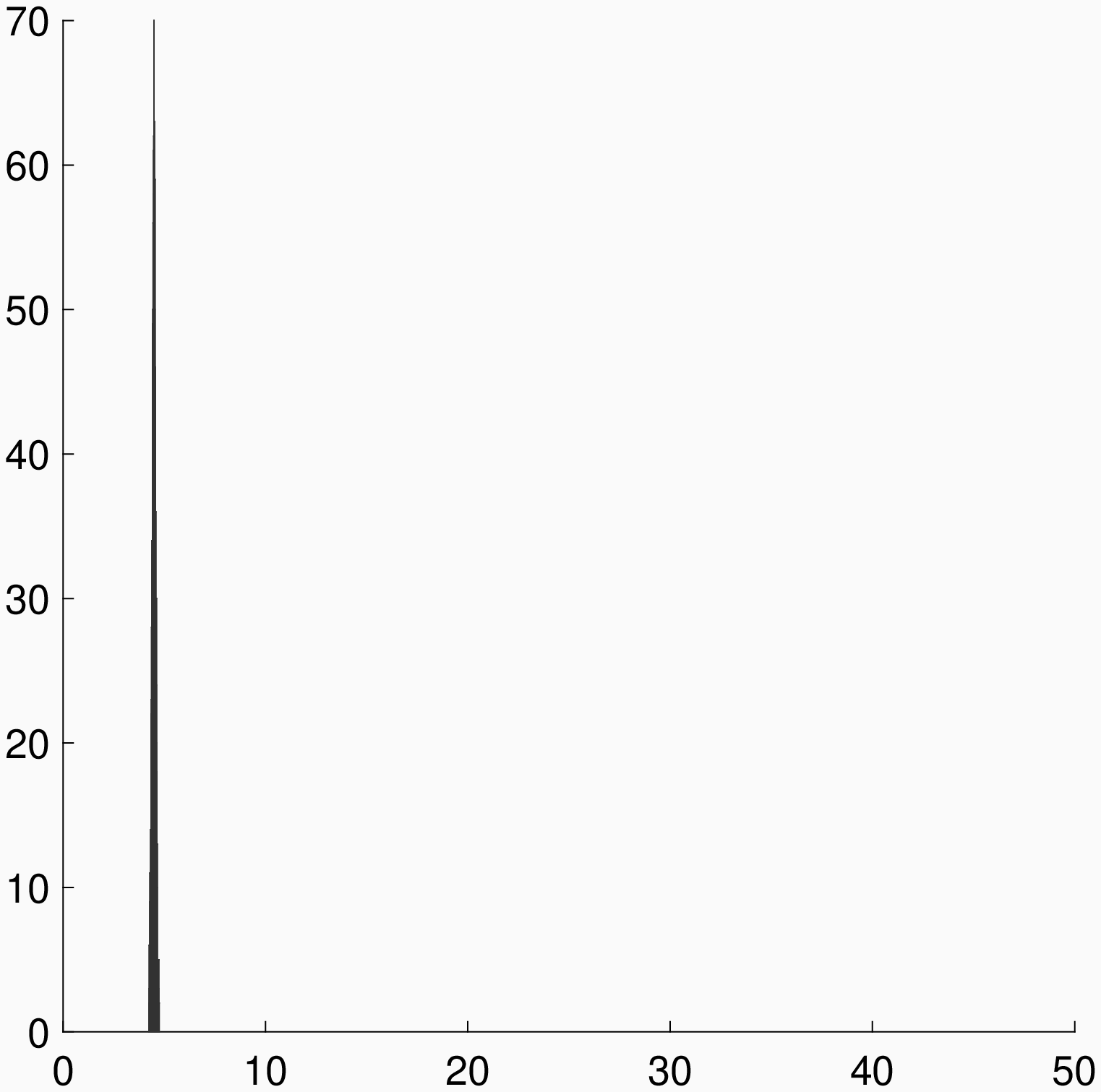}}
\put(326,125){\includegraphics[width=4.25cm]{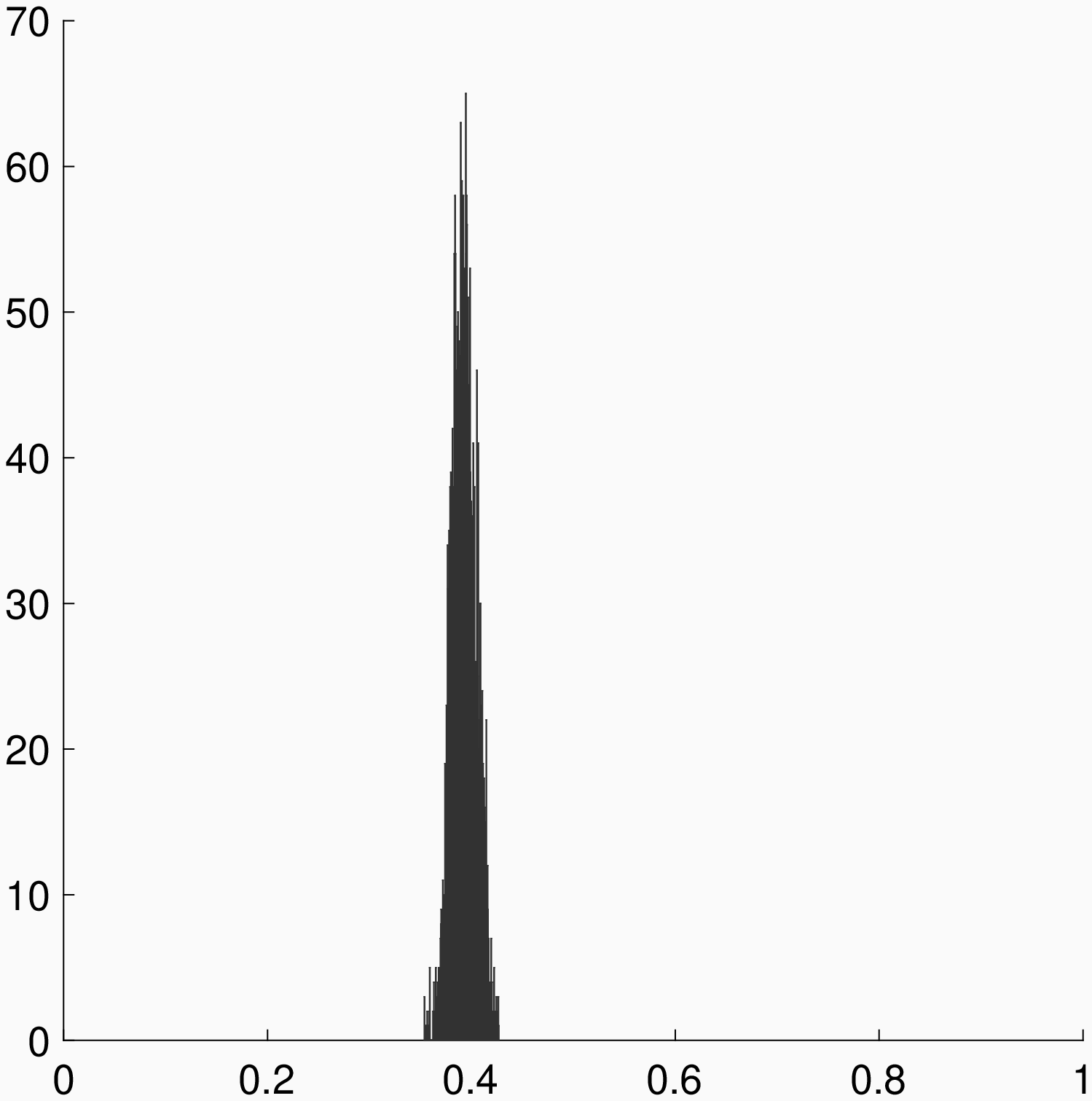}}
\put(52,-10){\includegraphics[width=4.25cm]{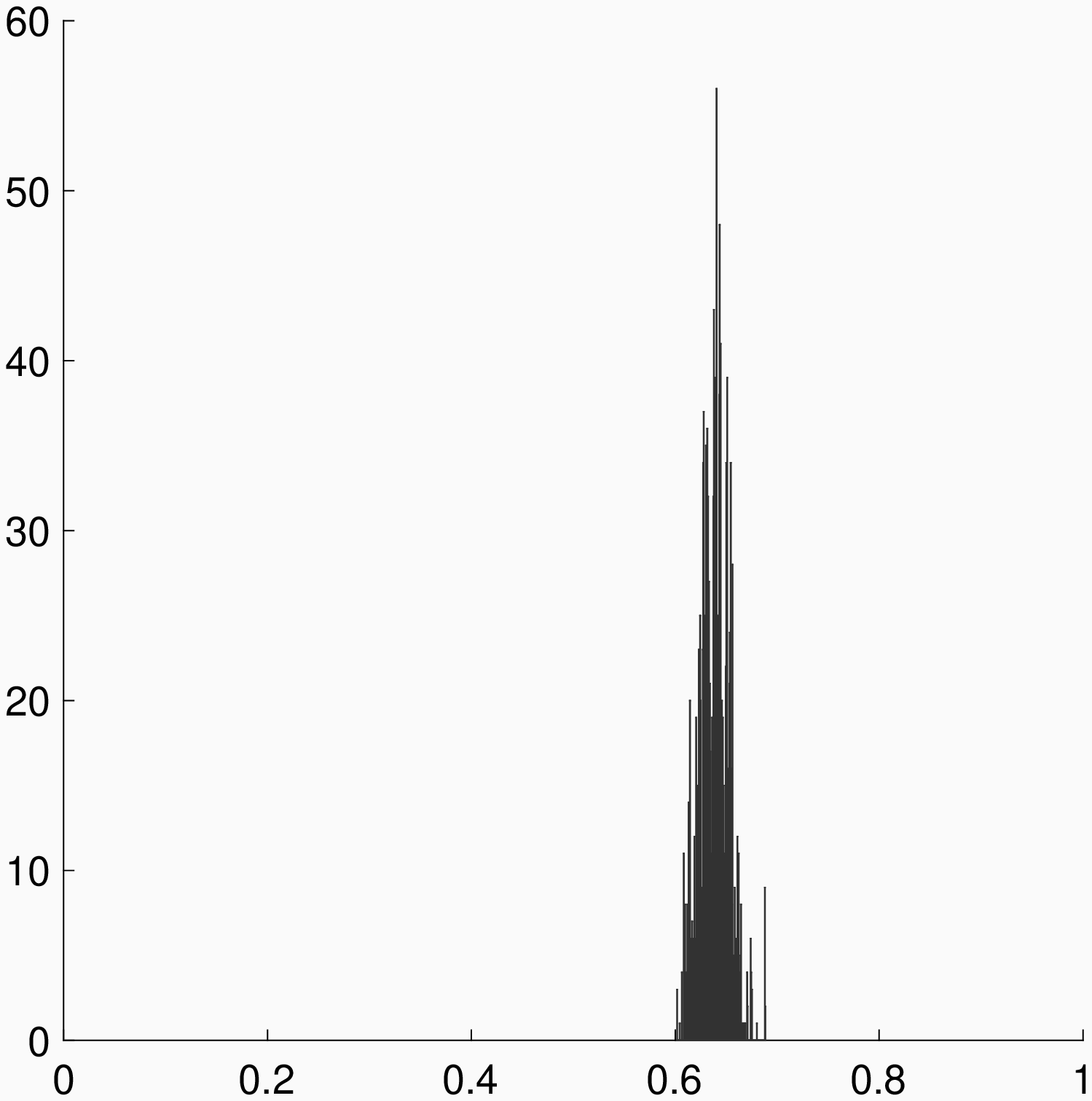}}
\put(326,-10){\includegraphics[width=4.25cm]{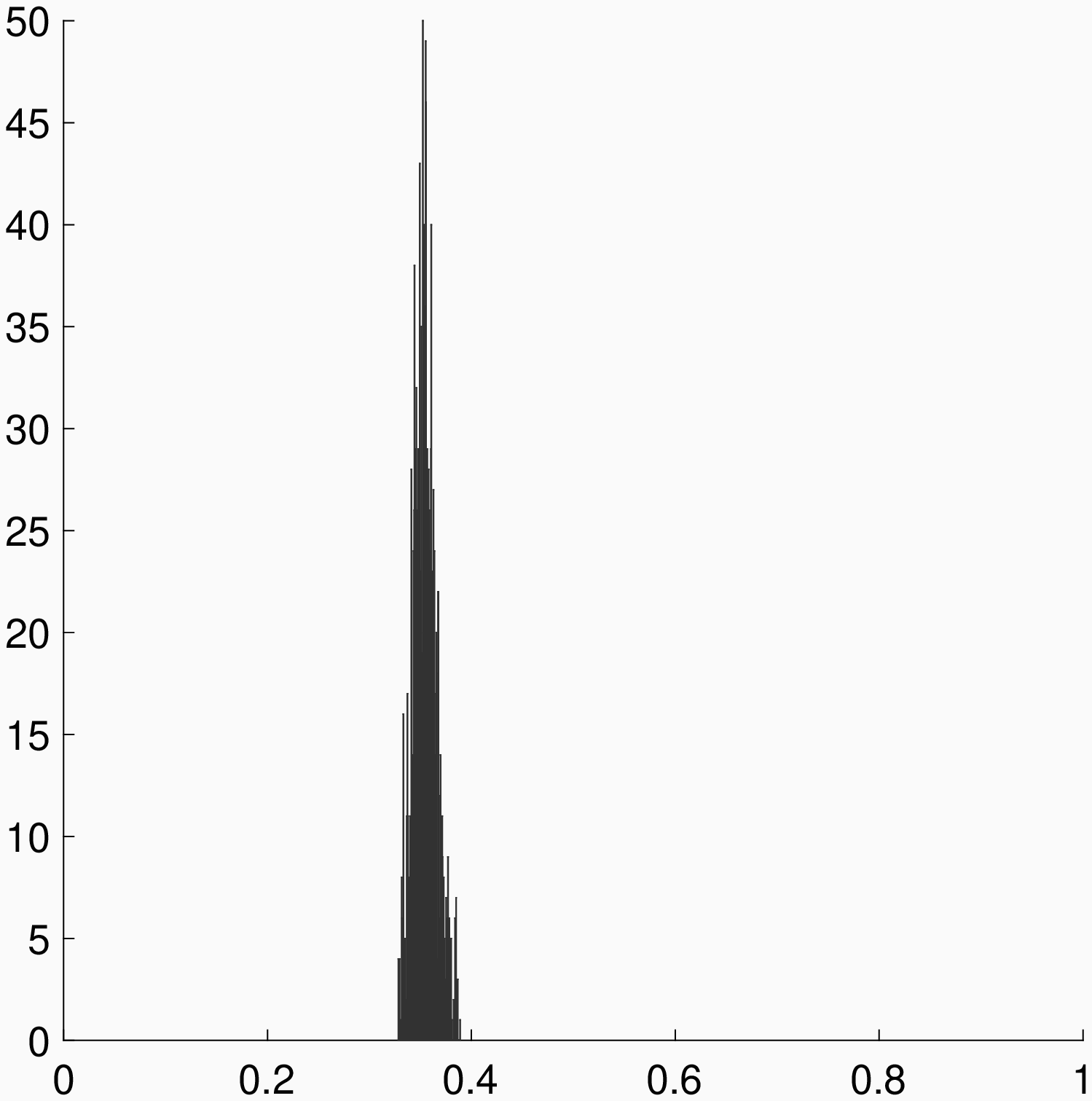}}
\put(30,480){\begin{rotate}{90} SE \end{rotate}}
\put(30,330){\begin{rotate}{90} Mat\'ern \end{rotate}}
\put(30,180){\begin{rotate}{90} Laplacian \end{rotate}}
\put(30,30){\begin{rotate}{90} Tikhonov \end{rotate}}
\put(125,560){$\sigma_f$}
\put(250,560){$l$}
\put(393,560){$\sigma$}
\end{picture}
\caption{Histogram of the 1-d marginal distribution of the GP parameters. Left, middle and right columns are the marginal distribution for parameter $\sigma_f$, $l$ and $\sigma$ with corresponding covariance functions indicated in the vertical text in the left of the figure. The estimate of the parameter $l$ is not available for Tikhonov covariance.}
\label{fig:Histogram parameters Chest}
\end{figure}

Several reconstructions of the chest phantom using different covariance functions, namely squared exponential (SE), Mat\'ern, Laplacian, and Tikhonov, are presented. For the SE, Mat\'ern, and Laplacian covariance functions, the paramameters $\sigma_f$, $l$, and $\sigma$ are estimated using the proposed method. We use $\nu = 1$ for the Mat\'ern covariance. As for the Tikhonov covariance, it is not characterized by the length scale $l$, and hence only $\sigma_f$ and $\sigma$ are estimated. All the estimated parameters are reported in Table~\ref{GP parameters}. Figure~\ref{fig:Histogram parameters Chest} presents the histograms of the 1-d marginal posterior distribution of each parameters using different covariance functions. The histograms show the distribution of the parameters samples in the Metropolis--Hastings samples. The results show that the $\sigma_f$ estimate for SE and Mat\'ern covariances is $0.12$, while for Laplacian and Tikhonov, the estimates are $0.05$ and $0.64$. For Mat\'ern, Laplacian, and Tikhonov covariance functions, the $\sigma$ estimates are concentrated around the same values $0.34-0.39$ with standard deviation (SD) between $0.02 - 0.03$. These noise estimates are well-estimated the ground-truth noise, $\sigma_{\text{true}} = 0.32$, with the absolute error is between $0.02 - 0.07$. The estimate of the SE kernel appears to overestimate the noise, $\sigma = 0.60$. It is reported that the length-scale parameter, $l$, for Laplacian and SE covariance functions are concentrated in the same values, while for Mat\'ern yields higher estimate, $l = 10.14$.

Figure~\ref{fig:ChestPhantomRec}(c)-(f) shows GP reconstructions of the 2D chest phantom using different covariance functions from 9 projections (uniformly spaced) out of 180$^\circ$ angle of view and $185$ number of rays for each projection.  The computation times for all numerical tests are reported in Table~\ref{Computation time}. The Metropolis--Hastings reconstruction shows longer computational time due to the need for generation of a large number of samples from the posterior distribution. However, the benefit of this algorithm is that it is easy to implement and it is reliable for sampling from high dimensional distributions.

\begin{table}[h]
\caption {Computation times of chest phantom (in seconds)} \label{Computation time}
\begin{center} 
\begin{tabular}{ | c | c |  c | c | c | c | } \hline 
Target &  FBP & SE & Mat\'ern & Laplacian & Tikhonov     \\ \hline 
\hline
Chest phantom   & 0.5 & 11\,210 & 9\,676 & 9\,615& 9\,615 \\
\hline
\end{tabular} 
\end{center}
\end{table}

The numerical test of the simulated data reconstructions is compared against figures of merit, namely:
\begin{itemize}
    \item the relative error (RE)
    \begin{align*}
        \frac{\|f_{\text{true}} - f_{\text{rec}} \|_2}{\| f_{\text{true}}\|_2},
    \end{align*}
    where $f_{\text{rec}}$ is the image reconstruction, and
    \item the peak-signal-to-noise ratio (PSNR)
     \begin{align*}
  10\log_{10}\left(\frac{\mathrm{peakval}^2}{\mathrm{MSE}}\right),
    \end{align*}
    where $\mathrm{peakval}$ is the maximum possible value of the image and $\mathrm{MSE}$ is the mean square error between $f_{\text{true}}$ and $f_{\text{rec}}$,
\end{itemize}
as shown in Table~\ref{Figures of merit}. 

In practice, image quality in CT depends on other parameters as well, such as image contrast, spatial resolution, and image noise \cite{goldman2007principles}. These parameters can be evaluated when the CT device is equipped with CT numbers for various materials, high-resolution image is available, and statistical fluctuations of image noise which require several times of measurement to record random variations in detected x-ray intensity are acquired. However, in this work, the collected datasets are not supported by the aforementioned factors and they fall outside the scope of this paper. The results presented here are focusing on the implementation of a new algorithm to limited-data CT reconstruction and are reported as a preliminary study.

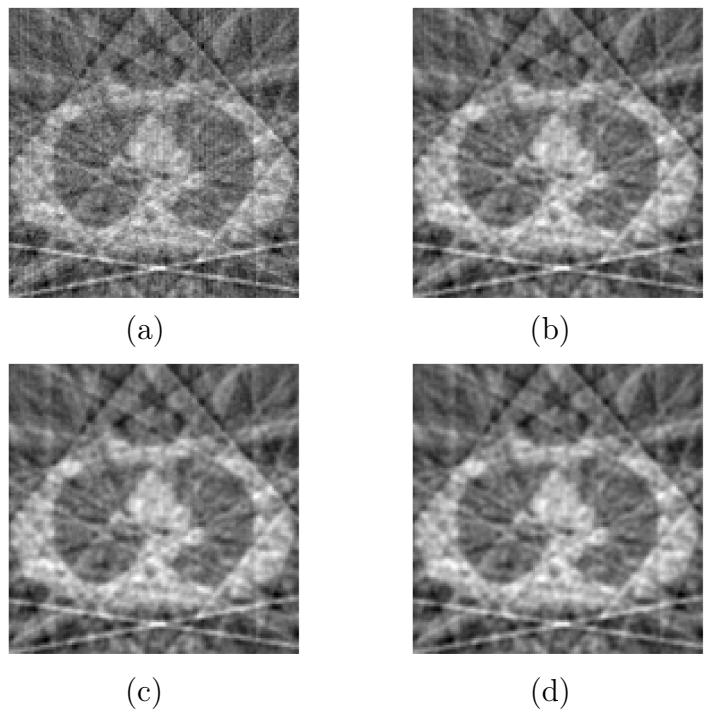
\begin{figure}
\centering
\begin{picture}(270,260)
\put(5,138){\includegraphics[width=6.3cm]{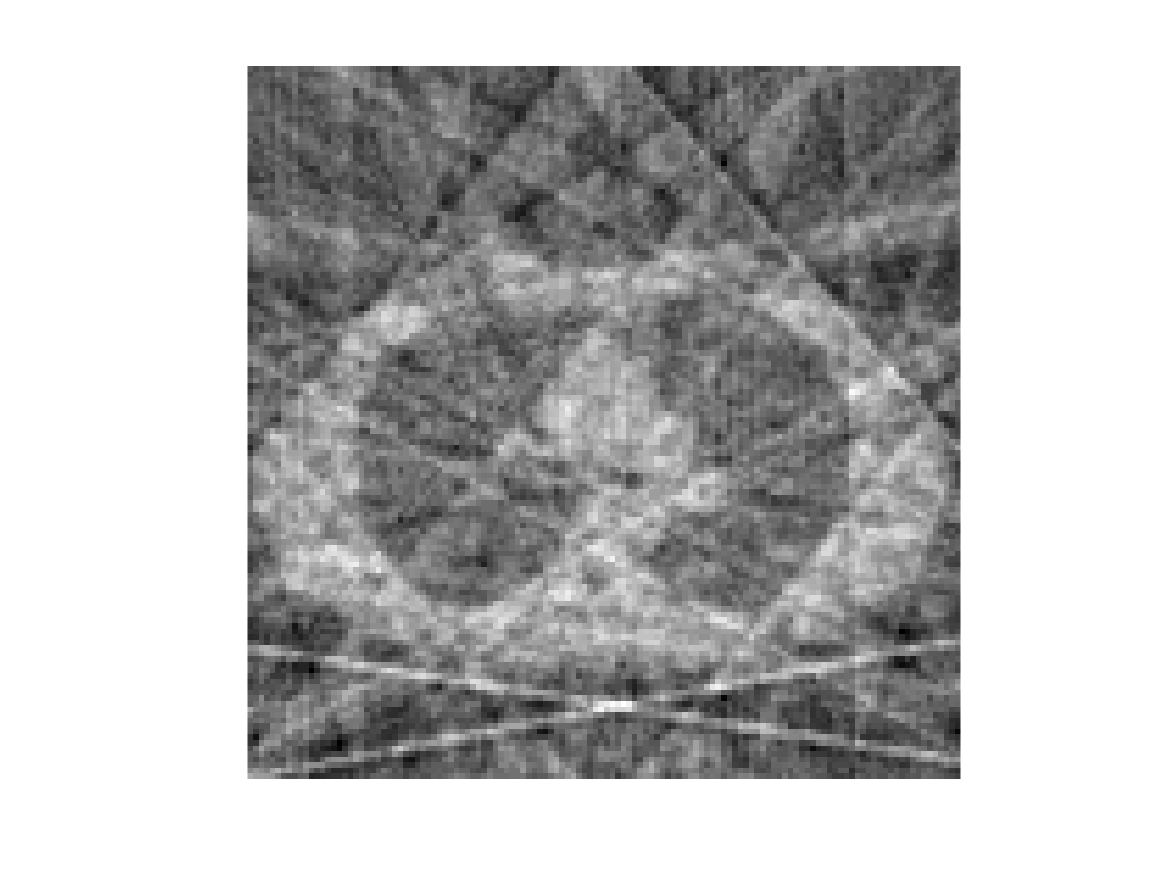}}
\put(158,138){\includegraphics[width=6.3cm]{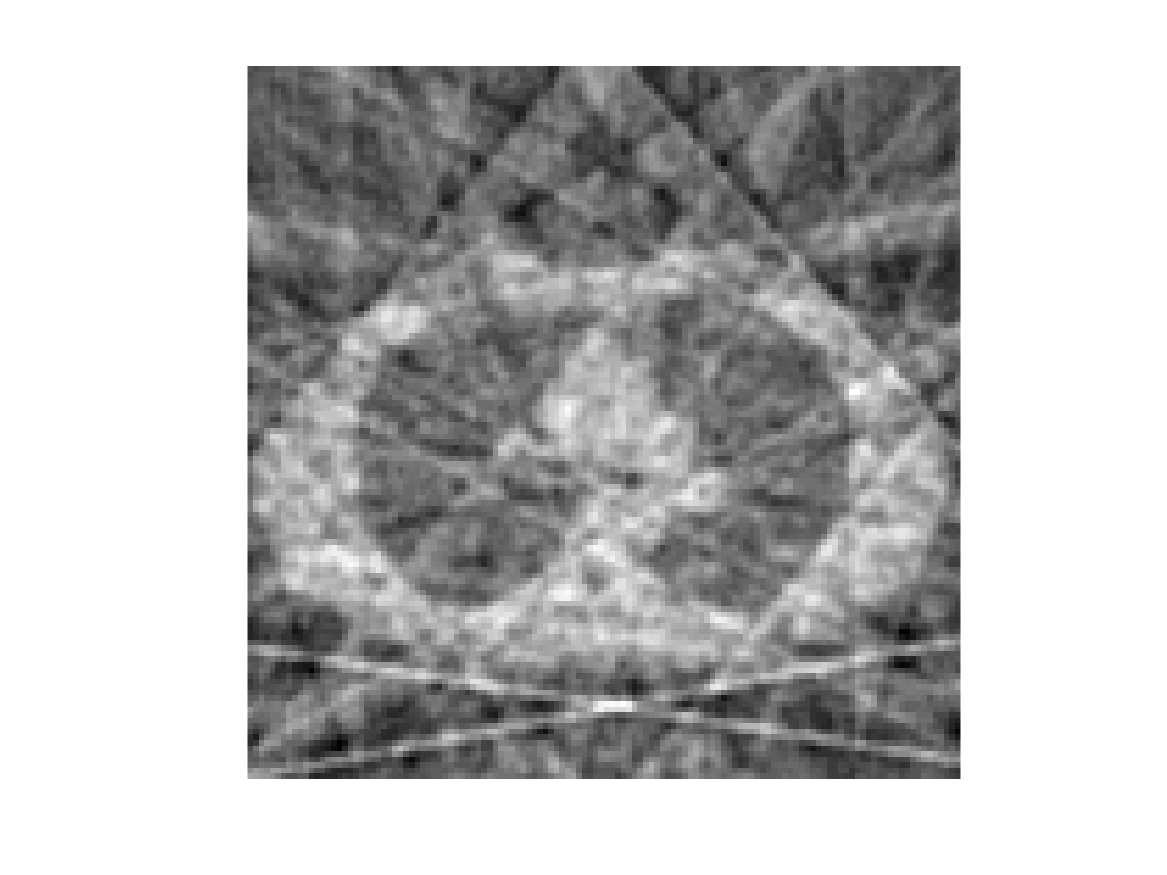}}
\put(5,3){\includegraphics[width=6.3cm]{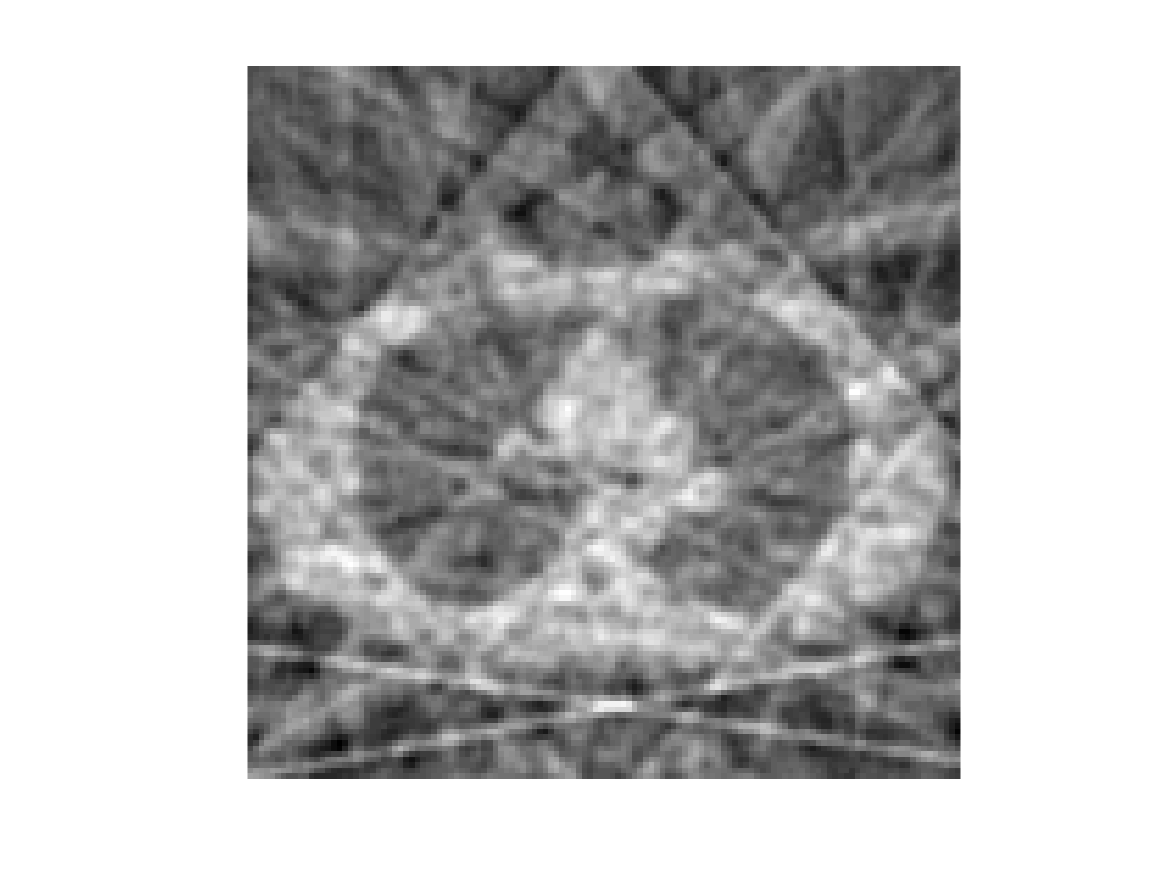}}
\put(158,3){\includegraphics[width=6.3cm]{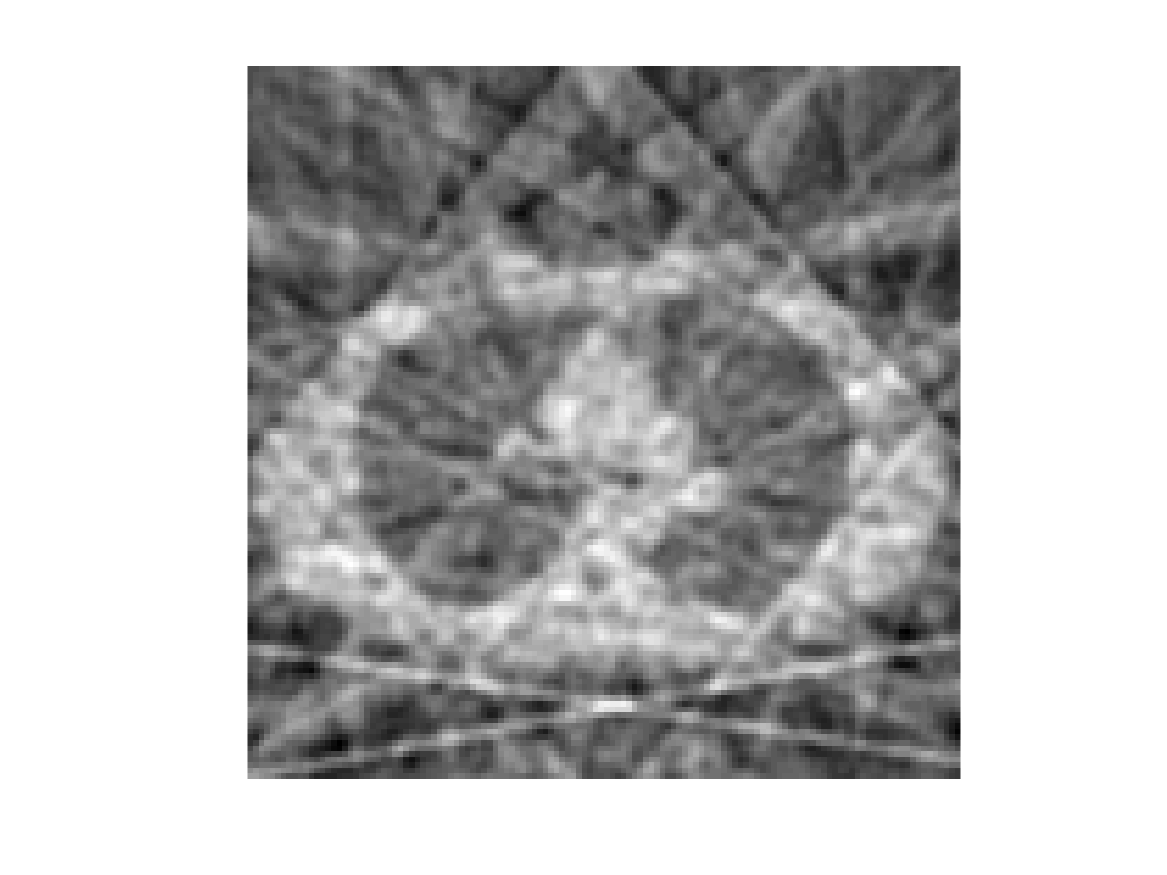}}
\put(87,137){(a)}
\put(240,137){(b)}
\put(87,0){(c)}
\put(240,0){(d)}
\end{picture}
\caption{Filtered backprojection reconstructions using (a) Shepp--Logan filter, (b) Cosine filter (c) Hamming filter, (d) Hann filter. Values of relative error (RE) are between $23.6 - 25.2$ and PSNR values are between $18.1 - 19.9\%$. }
\label{fig:FBPs}
\end{figure}

Reconstruction using a conventional method is computed as well with the built-in \textsc{Matlab} function {\tt iradon}, which uses the FBP to invert the Radon transform. It reconstructs a two-dimensional slice of the sample from the corresponding projections. The angles for which the projections are available are given as an argument to the function. Linear interpolation is applied during the backprojection and a Ram--Lak or ramp filter is used. The FBP reconstruction of the chest phantom is shown in Figure~\ref{fig:ChestPhantomRec}(b). For comparison, FBP reconstructions computed using some other filters are seen in Figure~\ref{fig:FBPs}.

\begin{table}[h]
\caption {The GP parameter estimates for the chest phantom. The estimates are calculated using the conditional mean, and the standard deviation (SD) values are also reported in parentheses.} \label{GP parameters}
\begin{center} 
\begin{tabular}{| c |  c | c | c | } \hline Covariance & $\sigma_f$ (SD) &  $l$ (SD)& $\sigma$ (SD) \\ function &  &   &  \\
\hline 
\hline
SE & 0.12 (0.04) & 5.03 (0.03) &  0.60 (0.02)\\
Mat\'ern & 0.12 (0.07)  & 10.14 (0.08)  & 0.34 (0.03) \\
Laplacian  & 0.05 (0.10)  & 4.49 (0.02) & 0.39 (0.03)\\
Tikhonov& 0.64  (0.02) & - & 0.35 (0.03)\\
\hline
\end{tabular} 
\end{center}
\end{table}

\begin{figure}
\begin{picture}(100,270)
\put(7,138){\includegraphics[width=6.3cm]{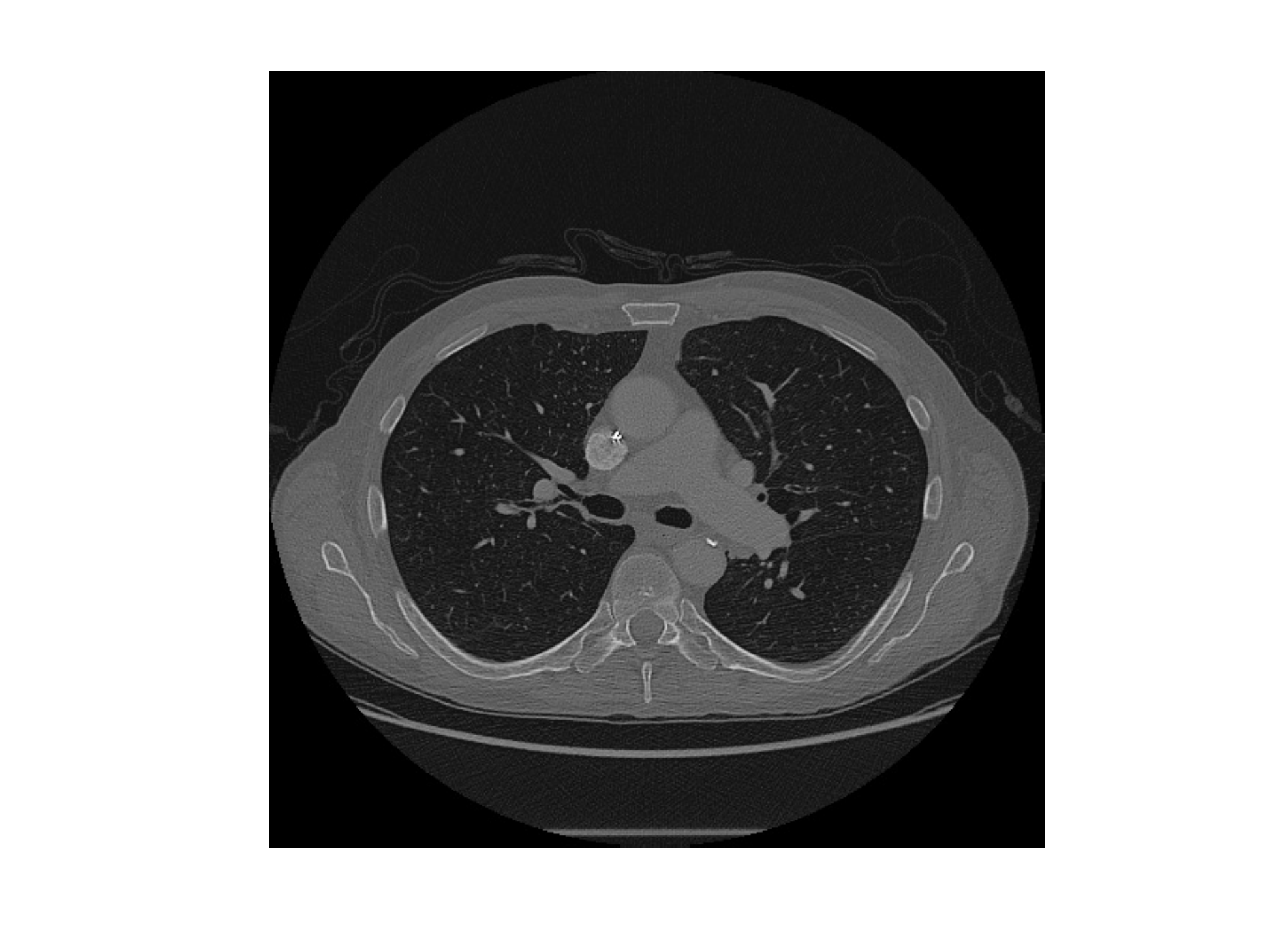}}
\put(155,138){\includegraphics[width=6.3cm]{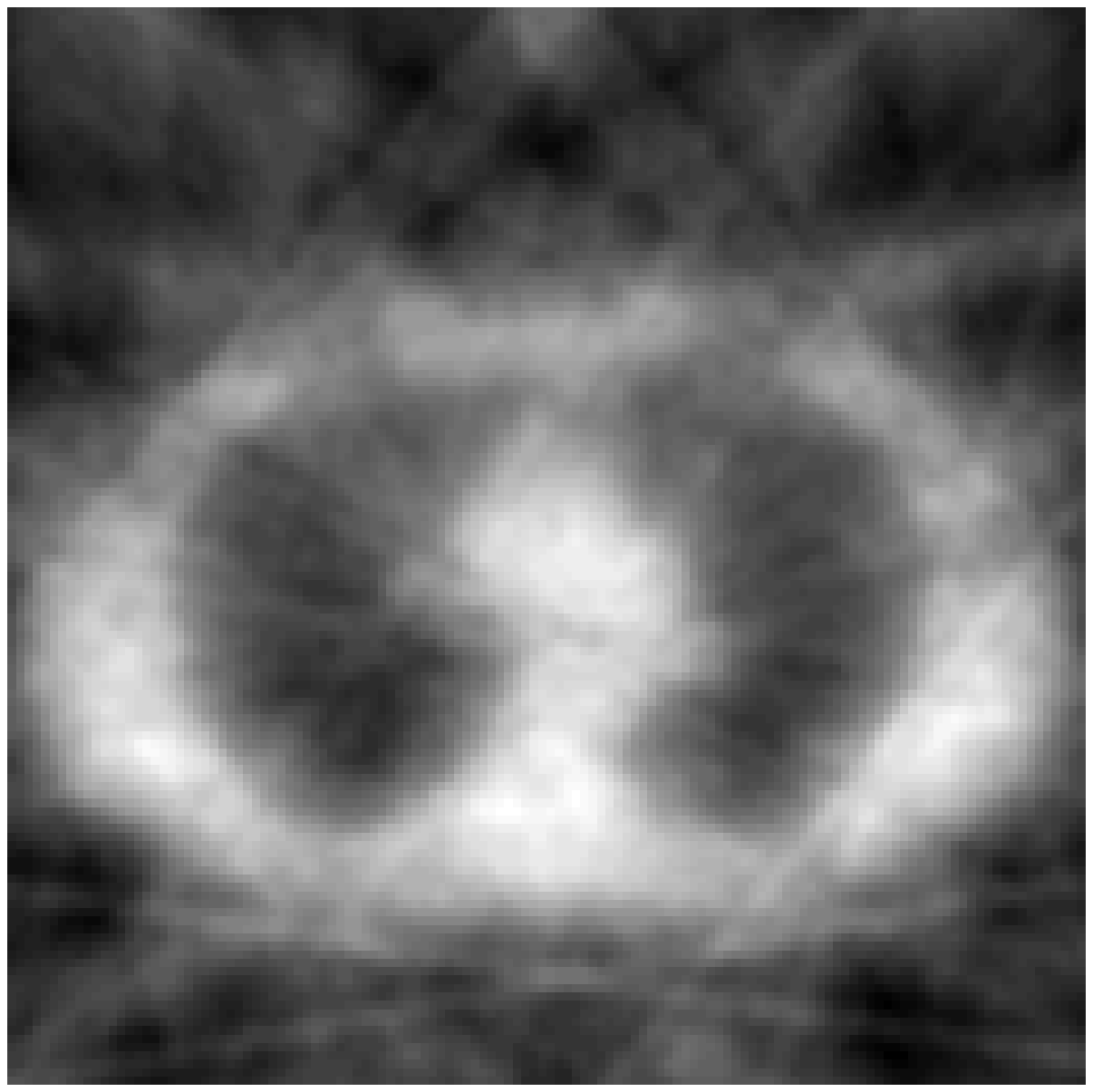}}
\put(306,138){\includegraphics[width=6.3cm]{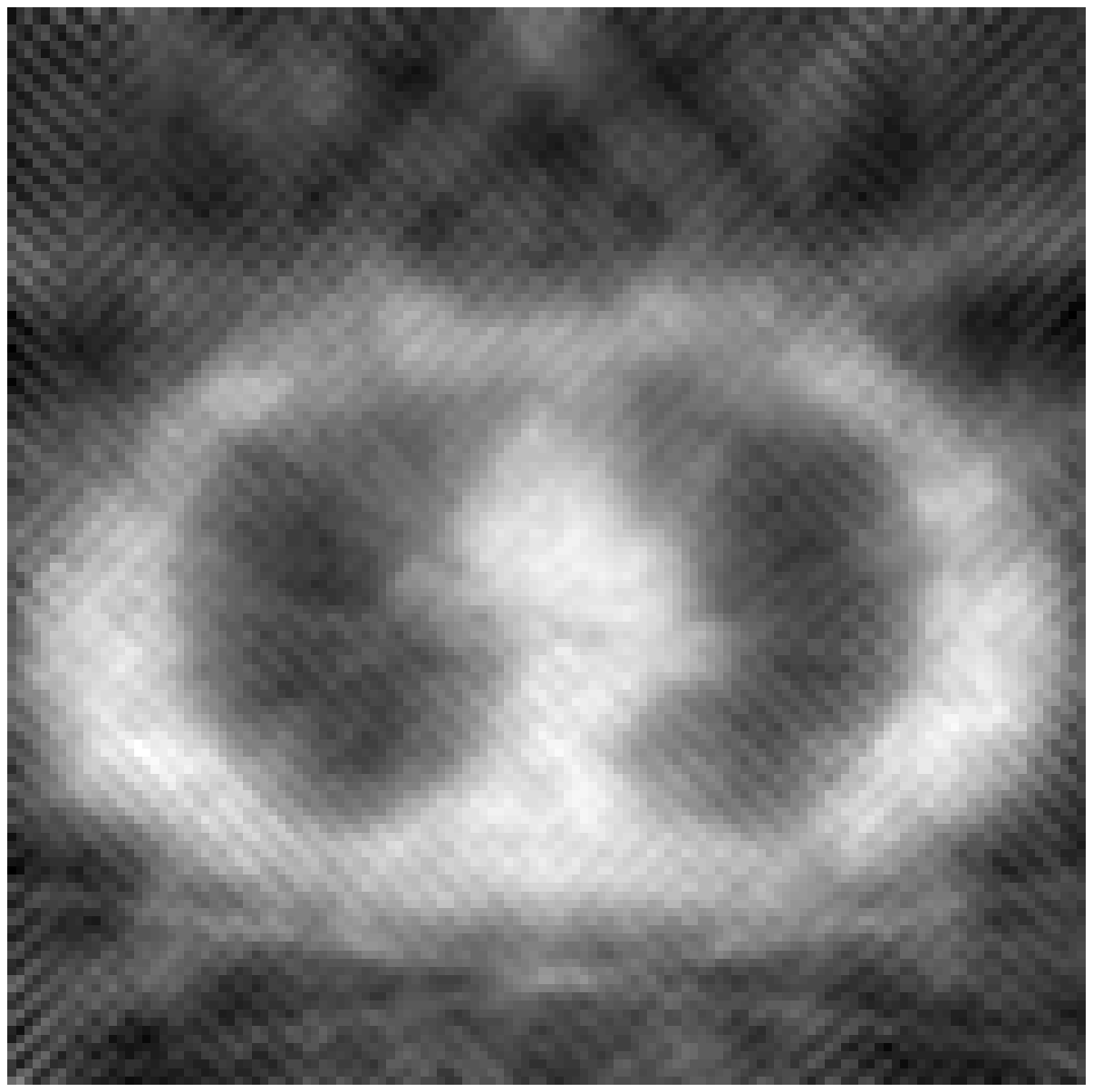}}
\put(8,3){\includegraphics[width=6.3cm]{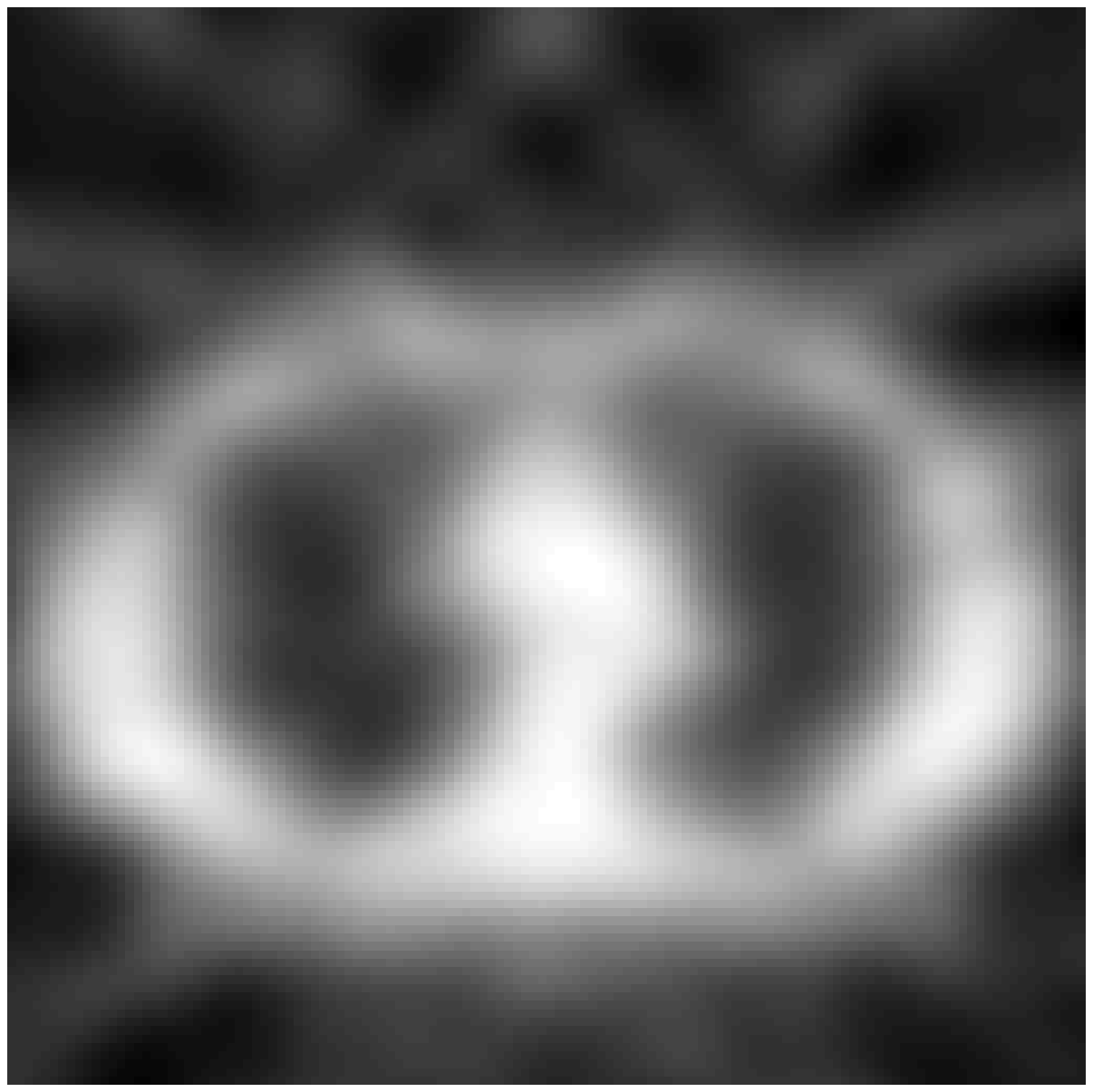}}
\put(155,3){\includegraphics[width=6.3cm]{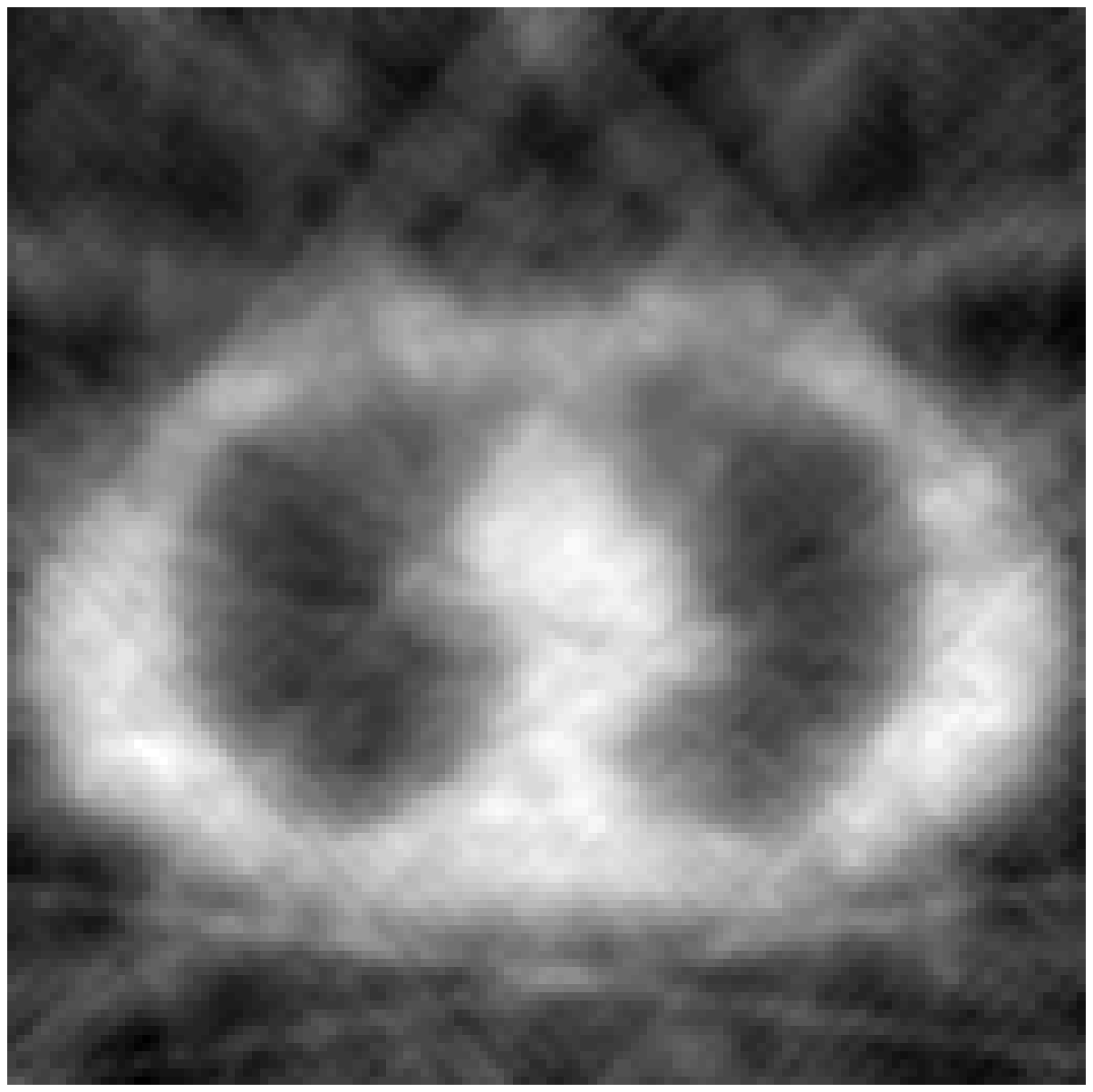}}
\put(87,137){(a)}
\put(240,137){(b)}
\put(393,137){(c)}
\put(87,0){(d)}
\put(240,0){(e)}
\end{picture}
\caption{(a) A ground truth of 2D chest phantom. (b) \& (c) reconstructions using L-curve parameter choice method with Laplacian (using $\sigma = 1$) and Tikhonov (using $\sigma = 0.2$) covariance functions, respectively. (d) \& (e) reconstructions using CV with Laplacian and Tikhonov covariance functions, respectively }
\label{fig:Parameter Choice Methods}
\end{figure}

\begin{table}[h]
\caption {Figures of merit for chest phantom reconstructions.} \label{Figures of merit}
\begin{center} 
\begin{tabular}{ | l | c |  c | c | } \hline Method &  RE (\%) & PSNR  \\ \hline 
\hline
FBP (Ram--Lak filter)  & 25.86 & 18.44 \\
GP-SE & 29.41 & 21.76 \\
GP-Mat\'ern  & 23.26 & 22.76 \\
GP-Laplacian   & 29.18 & 21.79 \\
GP-Tikhonov   & 23.39  & 22.73  \\
Lcurve-Laplacian  & 23.38 & 22.62\\
Lcurve-Tikhonov   & 23.26  & 22.63 \\
CV-Laplacian   & 25.18 & 22.31 \\
CV-Tikhonov   & 23.47 & 22.75 \\
\hline
\end{tabular} 
\end{center}
\end{table}
We also compared the results to the L-curve method and the CV:
\begin{itemize}
\item The L-curve method is applied to the Laplacian and the Tikhonov covariances and the L-curve plots from different values of parameter $10^{-1}\leq\sigma\leq10$ for both covariances are shown in Figure~\ref{L-curve chest data}. Both plots show that the corner of the L-curve is located in between $0.2\leq \sigma \leq 1$.
\begin{figure}[h]
\begin{picture}(100,250)
\put(-30,50){\includegraphics[width=9cm]{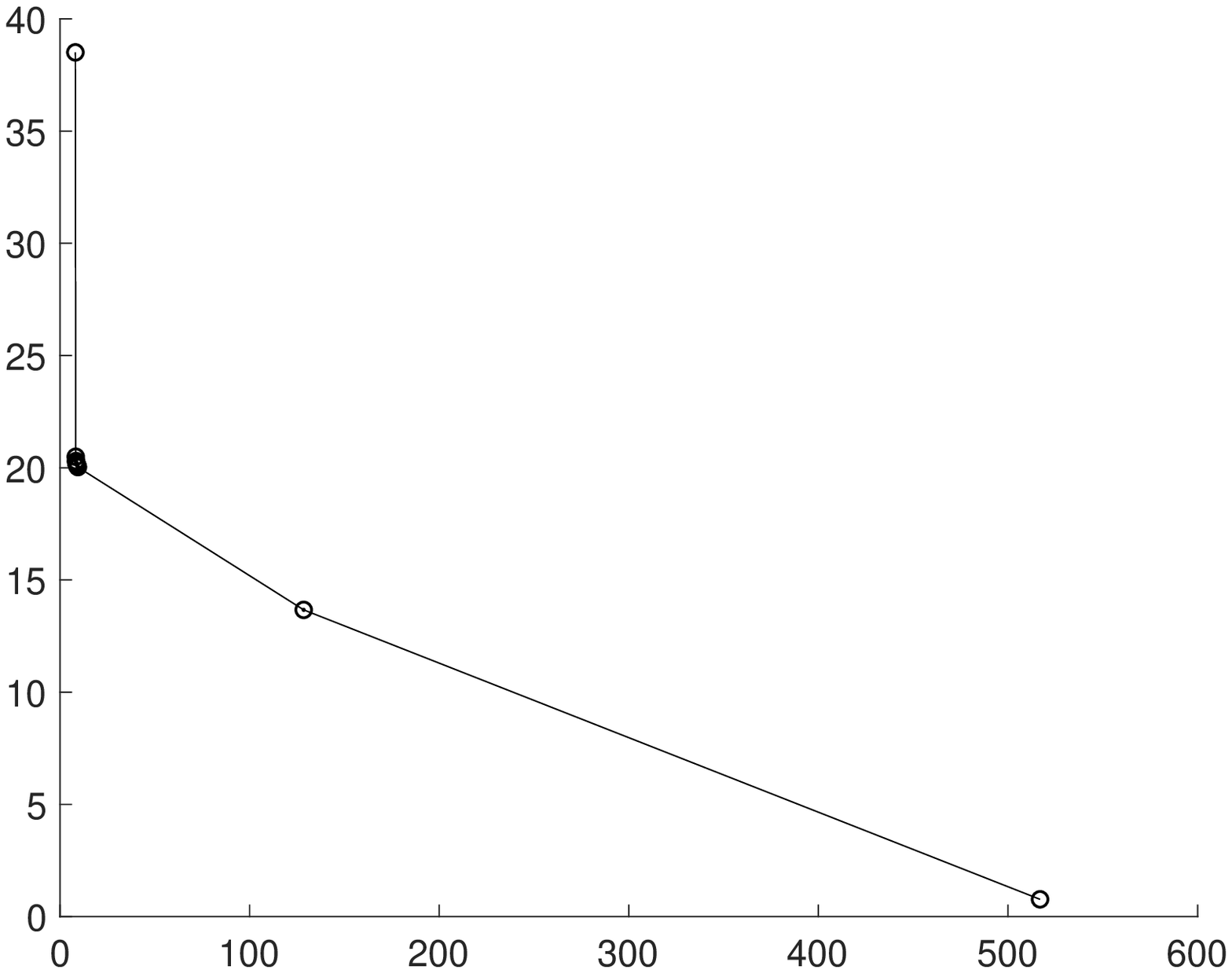}}
\put(210,50){\includegraphics[width=9cm]{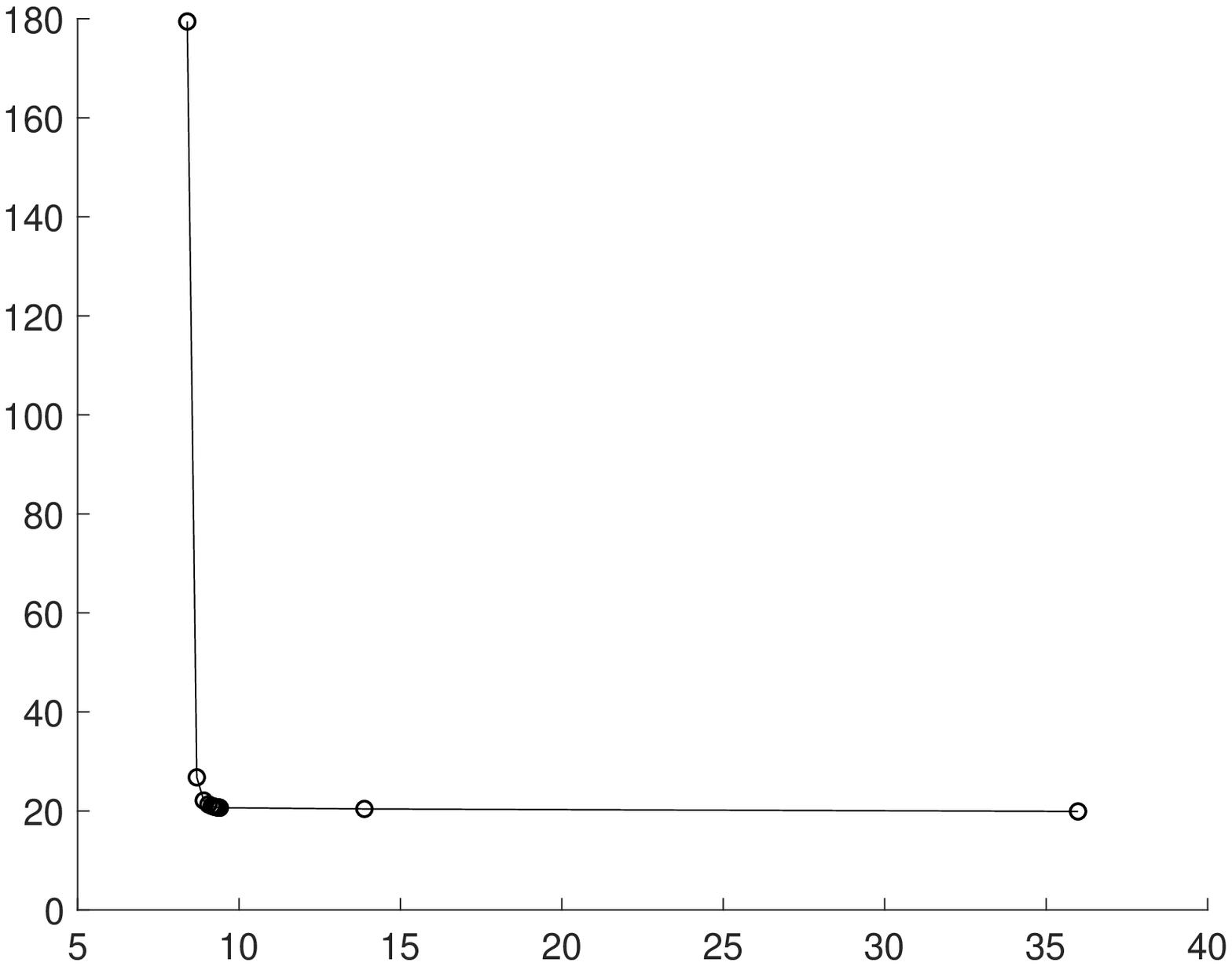}}
\put(65,45){$\lVert \mathcal{H}_{\x,i} f_{\sigma}(\x) - y_i \rVert _2$}
\put(-13,130){\begin{rotate}{90} $\lVert f_{\sigma}(\x) \rVert _2$\end{rotate}}
\put(225,130){\begin{rotate}{90} $\lVert f_{\sigma}(\x) \rVert _2$\end{rotate}}
\put(305,45){$\lVert \mathcal{H}_{\x,i} f_{\sigma}(\x) - y_i \rVert _2$}
\put(100,15){(a)}
\put(345,15){(b)}
\put(10,220){\small{$\sigma = 0.1$}}
\put(10,150){\small{$0.2\leq \sigma \leq 1$}}
\put(50,127){\small{$\sigma = 10$}}
\put(175,80){\small{$\sigma = 10^2$}}
\put(267,225){\small{$\sigma = 0.1$}}
\put(269,94){\rotatebox{45}{\small{$0.2\leq \sigma \leq 1$}}}
\put(290,95){\small{$\sigma = 10$}}
\put(410,95){\small{$\sigma = 10^2$}}
\end{picture}
\caption{The L-curve for (a) Tikhonov and (b) Laplacian covariance  from the chest phantom reconstruction.}\label{L-curve chest data}
\end{figure}

\item The CV is tested for the Laplacian and Tikhonov covariances using point-wise evaluation of  $10^{-2} \leq \sigma \leq 1$ and $10^{-2}\leq \sigma_f \leq 1$.
For the Laplacian covariance, several points of length scale $1\leq \ell \leq 100$ are tested as well. The minimum prediction error was obtained for $\sigma_f = 0.8$, $\sigma = 0.8$ and $\ell = 10$.
For the Tikhonov covariance, the minimum prediction error was obtained for $\sigma = 0.5$ and $\sigma_f = 0.5$. The estimates of $\sigma_f$ and $\sigma$ for Laplacian are $0.8$ and $0.5$, respectively, and they give the same estimates for the Tikhonov covariance function. The estimates of $\sigma$ for both kernels appear to overestimate the $\sigma_{\text{true}}$. The absolute error is in between $0.18 - 0.48$. The length-scale estimate from Laplacian covariance, $l = 10$, appears to close to the estimate in Mat\'ern covariance.

\end{itemize}
Image reconstructions for both L-curve and CV methods are shown in Figure~\ref{fig:Parameter Choice Methods}.

\subsection{Real data: Carved cheese}\label{RealData}

We now consider a real-world example using the tomographic x-ray data of a carved cheese slice measured with a custom-built CT device available at the University of Helsinki, Finland. The dataset is available online \cite{fips_dataset}. For a detailed documentation of the acquisition setup---including the specifications of the x-ray systems---see \cite{bubba2017tomographic}. We use the downsampled sinogram with $140$ rays and 15 projections from $360^\circ$ angle of view. In the computations, the size of the target is set to $120 \times 120$.

\begin{figure}
\begin{picture}(100,200)
\put(150,38){\includegraphics[width=6.21cm]{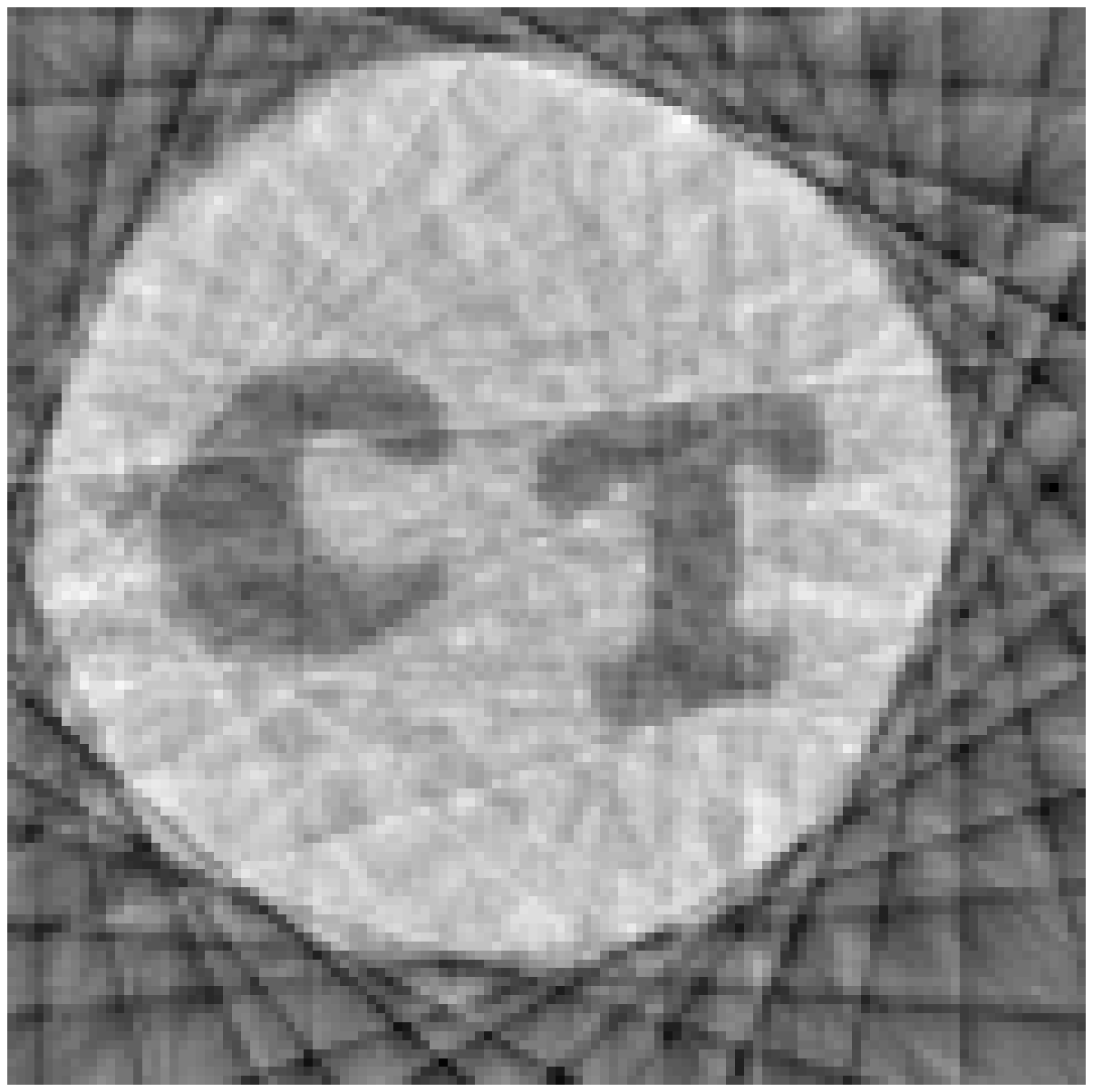}}
\put(0,-31){\includegraphics[width=6.8cm]{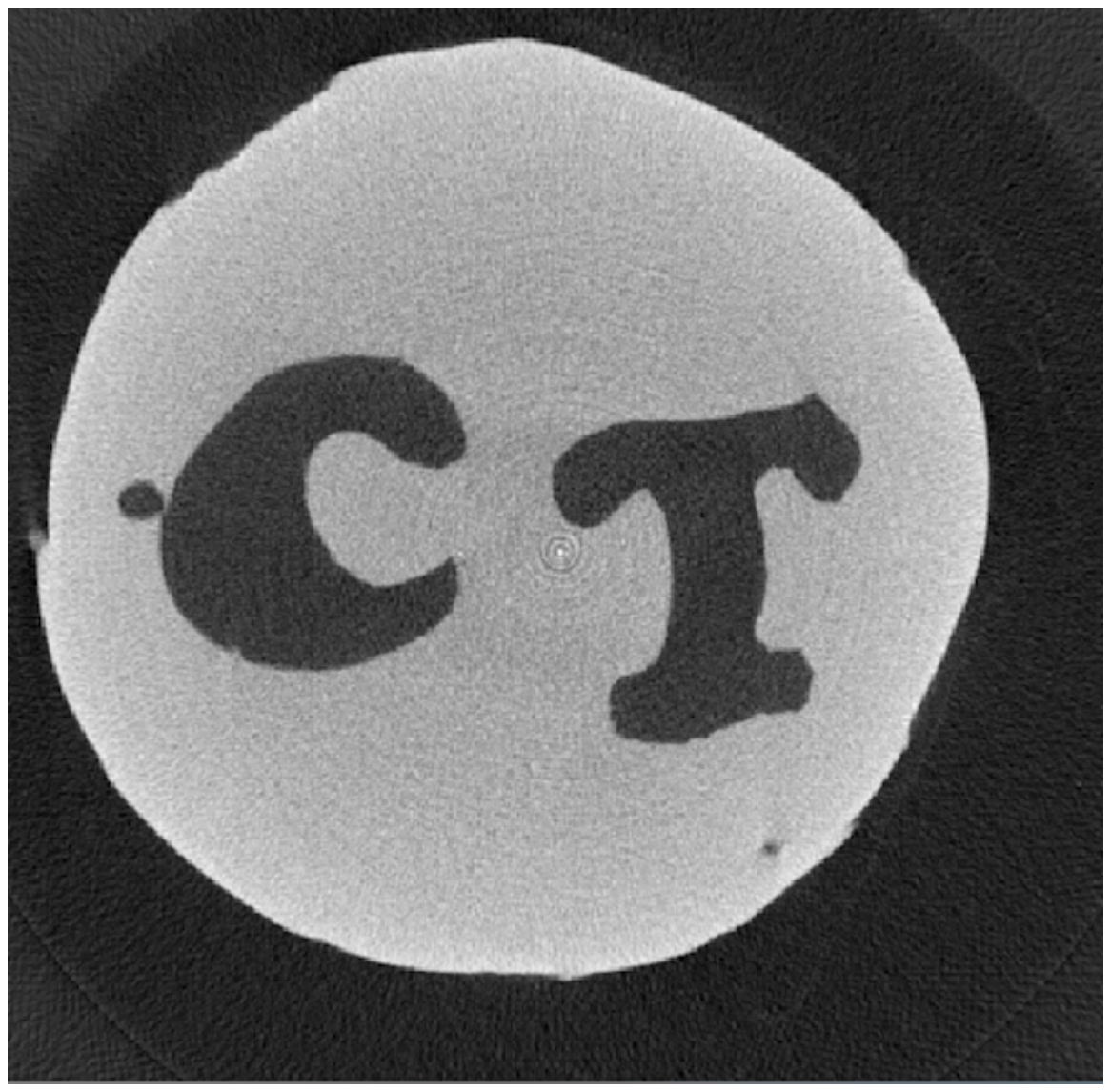}}
\put(295,38){\includegraphics[width=6.21cm]{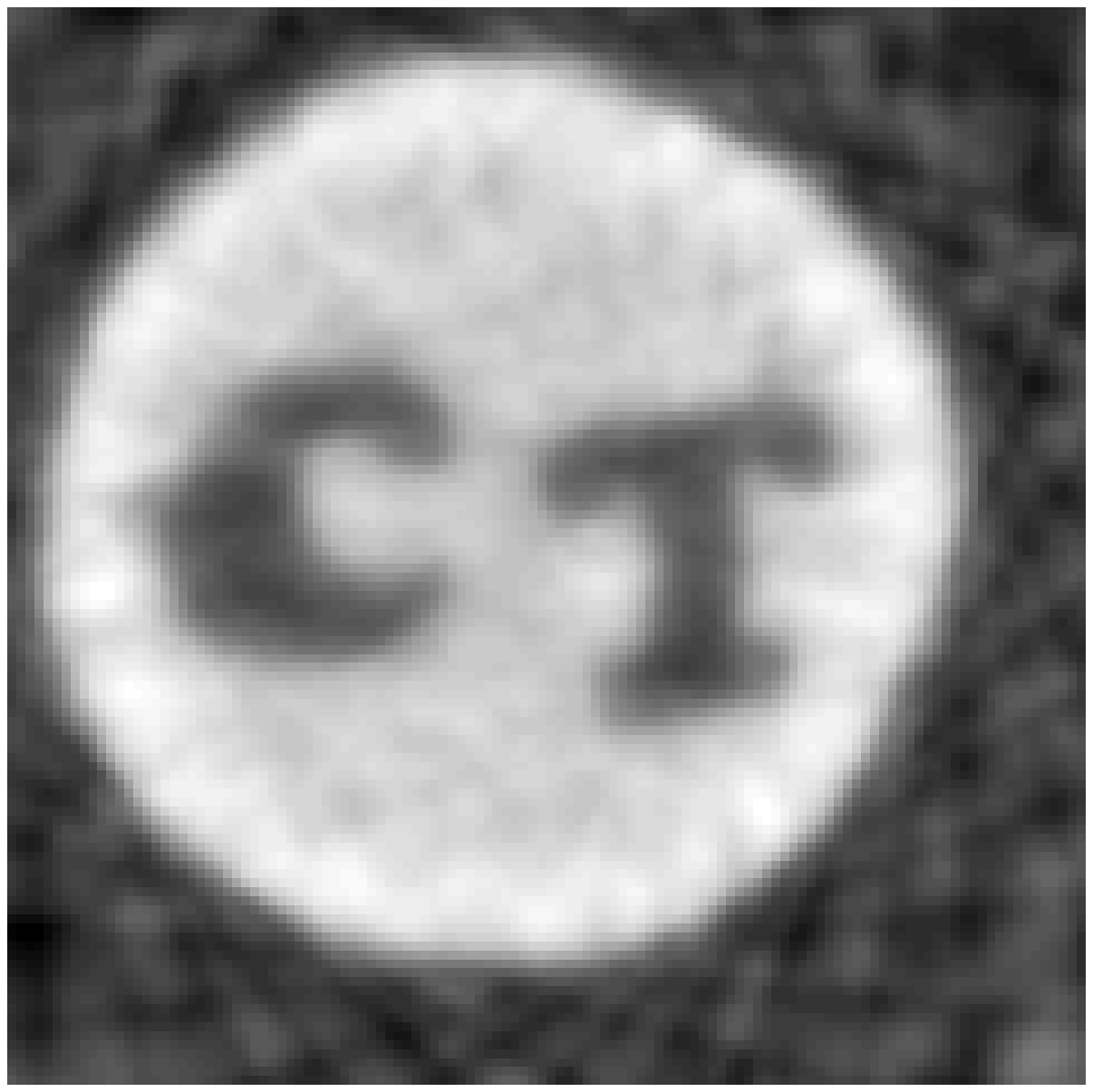}}
\put(90,30){(a)}
\put(232,30){(b)}
\put(395,30){(c)}
\end{picture}
\caption{(a) FBP reconstruction (Ram--Lak filter) of the carved cheese using dense $360$ projections. (b) Filtered backprojection reconstruction from 15 projections. (c) GP reconstruction using Mat\'ern covariance from 15 projections.}\label{CheeseRec}
\end{figure}

Figure~\ref{CheeseRec}(c) shows the GP reconstruction (Mat\'ern covariance function) of the cross section of the carved cheese slice using 15 projections (uniformly spaced) out of 360$^\circ$ angle of view. For comparison, the FBP reconstruction is shown in Figure~\ref{CheeseRec}(b).

\begin{table}[h]
\caption {Estimated GP parameters for the carved cheese using Mat\'ern covariance function. The estimates are calculated using the conditional mean, and the standard deviation (SD) values are also reported in parentheses.} \label{GP parameter cheese}
\begin{center} 
\begin{tabular}{| c | c |c |c |  } \hline Covariance & $\sigma_f$ (SD) & $l$ (SD) & $\sigma$ (SD) \\ 
 function   &  &  &\\ \hline 
\hline
Mat\'ern   & 0.012 (0.07) & 11.00 (0.08) & 0.02 (0.04) \\
\hline
\end{tabular} 
\end{center}
\end{table}

The computation times for the carved cheese are reported in Table~\ref{Computation time cheese}.

\begin{table}[h]
\caption {Computation times of the carved cheese (in seconds)} \label{Computation time cheese}
\begin{center} 
\begin{tabular}{ | c | c |  c | c | c | c | } \hline 
Target & FBP & Mat\'ern  \\ \hline 
\hline
Carved cheese & 0.1 & 12\,604\\
\hline
\end{tabular} 
\end{center}
\end{table}

\subsection{Discussion}
We have presented x-ray tomography reconstructions from both simulated and real data for limited projections (i.e. sparse sampling) using an approach based on the Gaussian process.  However, other limited-data problems such as limited angle tomography could be explored as well.
The quality of GP reconstructions using different covariance functions looks rather the same qualitatively. However, quantitatively, the reconstruction using Mat\'ern covariance is the best one: it has the lowest RE $23.26\%$ and the highest PSNR $22.76$. PSNR describes the similarity of the original target with the reconstructioned image (the higher value, the better of the reconstruction). Figures of merit estimates are not available for the real cheese data since there is no comparable ground truth. Nevertheless, the quality of the reconstruction can be observed qualitatively by comparing with the FBP reconstruction obtained with dense $360$ projections from $360$ degrees shown in  Figure~\ref{CheeseRec}(a). The corresponding parameter estimates for the chest phantom and the cheese are reported in Table~\ref{GP parameters} and \ref{GP parameter cheese}. For the chest phantom case, the estimate of parameter $\sigma$ using Mat\'ern, Laplacian and Tikhonov kernels tend to be close to the true value $\sigma_{true}$. As for the SE covariance, the standard deviation of noise is overestimated.

The reconstructions produced by the FBP benchmark algorithm using sparse projections are overwhelmed by streak artefacts due to the nature of backprojection reconstruction, as shown in Figure~\ref{fig:ChestPhantomRec}(b) for the chest phantom and Figure~\ref{CheeseRec}(b) the for cheese target. The edges of the target are badly reconstructed. Due to the artefacts, especially for the chest phantom, it is difficult to distinguish the lighter region (which is assumed to be tissue) and the black region (air). The FBP reconstruction has the worst quality and it is confirmed in Table~\ref{Figures of merit} that it has a high RE value ($25.86\%$) and the lowest PSNR ($18.44$). FBP reconstructions computed with different filters are shown in Figure~\ref{fig:FBPs}. However, there is no significant improvement in the images as it is clarified by the RE and PSNR values in the caption as well as by qualitative investigation. On the other hand, the GP reconstructions outperform the FBP algorithm in terms of image quality as reported in the figures of merit. The PSNR values of the GP-based reconstructions are higher than that of the FBP reconstruction. Nevertheless, in GP reconstructions, sharp boundaries are difficult to achieve due to the smoothness assumptions embedded in the model.

The GP prior clearly suppresses the artifacts in the reconstructions as shown in Figure~\ref{fig:ChestPhantomRec}(c) and \ref{CheeseRec}(c). In Figure~\ref{fig:ChestPhantomRec}(c), the air and tissue region are recovered much better, since the prominent artefacts are much less. In Figure~\ref{CheeseRec}(c), the air region (outside the cheese and the C and T letters) are much sharper than in the FBP reconstruction. Overall, the results indicate that the image quality can be improved significantly by employing the GP method.

In Figure~\ref{fig:Parameter Choice Methods} the image reconstructions using L-curve and CV methods are presented. The quality of the reconstructions is reported in Table~\ref{Figures of merit} as well. In these methods, finer point-wise evaluations might help to improve the quality of the reconstructions.

We emphasize that in the proposed GP-approach, some parameters in the prior is a part of the inference problem (see Equation \eqref{eq:SE}). Henceforth, we can avoid the difficulty in choosing the prior parameters. This problem corresponds to the classical regularization methods, in which selecting the regularization parameters is a very crucial step to produce a good reconstruction.

\section{Conclusions}
We have employed the Gaussian process with a hierarchical prior to computed tomography using limited projection data. The method was implemented to estimate the x-ray attenuation function from the measured data produced by the Radon transform. The performance has been tested on simulated and real data, with promising results shown. Unlike algorithms commonly used for the limited x-ray tomography problem that require manual tuning of prior parameters, the proposed GP method offers an easier set up as it takes into account the prior parameters as a part of the estimation. Henceforth, it constitutes a promising and user-friendly strategy. 

The most important part of the GP model is the selection of the covariance function, since it stipulates the properties of the unknown function. As such, it also leaves most room for improvement. Considering the examples in Section \ref{sec:expResults}, a common feature of the target functions is that they consist of a number of well-defined, separate regions. 
The function values are similar and thus highly correlated within the regions, while the correlation is low at the edges where rapid changes occur. This kind of behavior is hard to capture with a stationary covariance function that models the correlation as completely dependent on the distance between the input locations. A non-stationary alternative is provided by, for example, the neural network covariance function, which is known for its ability to model functions with non-smooth features \cite{Rasmussen2006}. The basis function approximation method employed in this work is only applicable to stationary covariance functions, but other approximations can of course be considered.

Despite its success, the computational burden of the proposed algorithm is relatively high. To solve this problem, speed-up strategies are available, such as implementing parallelized GPU code, optimizing the covariances of the sampling strategy, or by changing the MCMC algorithm to another one.
Investigating finer resolution images and statistical records would also be interesting future research to evaluate other image quality parameters. Moreover, the proposed method can be applied to multidetector CT imaging \cite{mookiah2018multidetector, flohr2005multi} as well as 3D CT problems using sparse data \cite{sidky2008image,purisha2018automatic}.

\ack{Authors acknowledge support from the Academy of Finland (314474 and 313708).
Furthermore, this research was financially supported by the Swedish Foundation for Strategic Research (SSF) via the project \emph{ASSEMBLE} (contract number: RIT15-0012).}

\appendix
\section{Details on the computation of $\Phi$}\label{app:compdet}
Here we derive the closed-form expression of the entries $\Phi_{ij}$ stated in \eqref{eq:Phi_entry}. 
We get that
\begin{equation}
\begin{split}
\Phi_{ij} &= \int_{-R}^R \phi_i(\x^0_j+s\hat{\vu}_j)ds \\
&=\frac{1}{\sqrt{L_1L_2}}\int_{-R}^{R} \sin(\varphi_{i_1} r_j\cos\theta_j - \varphi_{i_1} s \sin\theta_j +\varphi_{i_1}L_1)\sin(\varphi_{i_2} r_j\sin\theta_j + \varphi_{i_2} s \cos\theta_j+\varphi_{i_2} L_2) ds 
\\
& = \frac{1}{\sqrt{L_1L_2}}\int_{-R}^{R} \sin(\alpha_{ij} s + \beta_{ij})\sin(\gamma_{ij} s + \delta_{ij}) ds \\ 
& = \frac{1}{2\sqrt{L_1L_1}} \int_{-R}^R \cos((\alpha_{ij}-\gamma_{ij})s+\beta_{ij}-\delta_{ij}) - \cos((\alpha_{ij}+\gamma_{ij})s + \beta_{ij} + \delta_{ij})ds  \\
& = \frac{1}{2\sqrt{L_1L_1}} \Big[ \frac{1}{\alpha_{ij}-\gamma_{ij}}\sin((\alpha_{ij}-\gamma_{ij})s+\beta_{ij}-\delta_{ij}) - \frac{1}{\alpha_{ij}+\gamma_{ij}}\sin((\alpha_{ij}+\gamma_{ij})s + \beta_{ij} + \delta_{ij})\Big]_{-R}^R 
\\
& = \frac{1}{2\sqrt{L_1L_1}} \Big( \frac{1}{\alpha_{ij}-\gamma_{ij}}\sin((\alpha_{ij}-\gamma_{ij})R+\beta_{ij}-\delta_{ij}) - \frac{1}{\alpha_{ij}+\gamma_{ij}}\sin((\alpha_{ij}+\gamma_{ij})R + \beta_{ij} + \delta_{ij}) 
\\
& \qquad - \frac{1}{\alpha_{ij}-\gamma_{ij}}\sin(-(\alpha_{ij}-\gamma_{ij})R+\beta_{ij}-\delta_{ij}) + \frac{1}{\alpha_{ij}+\gamma_{ij}}\sin(-(\alpha+\gamma_{ij})R + \beta_{ij} + \delta_{ij})\Big),
\end{split}
\end{equation}
where 
\begin{subequations}
\begin{align}
	\alpha_{ij} &= \varphi_{i_1}\sin\theta_j, \\
    \beta_{ij}  &= \varphi_{i_1}r_j\cos\theta_j+\varphi_{i_1}L_1, \\
    \gamma_{ij} &= \varphi_{i_2}\cos\theta_j, \\
    \delta_{ij} &= \varphi_{i_2}r_j\sin\theta_j+\varphi_{i_2}L_2.
\end{align}
\end{subequations}

\section*{References}
\bibliographystyle{ieeetr}
\bibliography{sample}
\end{document}